\documentclass[10pt,twocolumn,letterpaper,table]{article}

\usepackage[pagenumbers]{cvpr} 

\usepackage[dvipsnames]{xcolor}

\definecolor{cvprblue}{rgb}{0.21,0.49,0.74}
\usepackage[pagebackref,breaklinks,colorlinks,citecolor=cvprblue]{hyperref}

\usepackage{amsmath}
\usepackage{color}
\usepackage{booktabs}
\usepackage{caption}
\usepackage{subcaption}
\usepackage{enumitem}
\usepackage{multirow}
\usepackage{multicol}
\usepackage{cuted}
\usepackage{float}
\usepackage{pifont}

\usepackage[toc,page,header]{appendix}
\usepackage{minitoc}

\def\paperID{3775} 
\def\confName{CVPR}
\def\confYear{2024}

\title{Snap Video: Scaled Spatiotemporal Transformers for Text-to-Video Synthesis}

\author{
{Willi Menapace\textsuperscript{1,2,*}
\quad
Aliaksandr Siarohin\textsuperscript{1}
\quad
Ivan Skorokhodov\textsuperscript{1}
\quad
Ekaterina Deyneka\textsuperscript{1}}\\\vspace{-1cm}
\and
{Tsai-Shien Chen\textsuperscript{1,3,*}
\quad
Anil Kag\textsuperscript{1}
\quad
Yuwei Fang\textsuperscript{1}
\quad
Aleksei Stoliar\textsuperscript{1}
\quad
Elisa Ricci\textsuperscript{2,4}}\\[-8mm]
\and
{Jian Ren\textsuperscript{1}
\quad
Sergey Tulyakov\textsuperscript{1}}\\[-4mm]\and
{Snap Inc.\textsuperscript{1}\quad University of Trento\textsuperscript{2}\quad UC Merced\textsuperscript{3}\quad Fondazione Bruno Kessler\textsuperscript{4}}\\[1mm]
{\href{https://snap-research.github.io/snapvideo/}{\small \url{snap-research.github.io/snapvideo}}}
}

\definecolor{darkorange}{rgb}{1.0, 0.55, 0.0}

\newcommand{\ivan}[1]{\textcolor{magenta}{}}
\newcommand{\jian}[1]{{\color{darkorange}{\textbf{}}}}

\newcommand{\sergey}[1]{{\color{cyan}{}}}
\newcommand{\mistery}[1]{{\color{purple}{\textbf{}}}}

\newcommand{\apref}[1]{Appx.~\ref*{#1}}
\newcommand{\website}{\emph{Website}}
\newcommand{\supp}{\emph{Appendix}}
\newcommand{\methodname}{Snap Video}

\DeclareMathOperator{\Var}{Var}

\newcommand{\titlerowww}[1]{\makebox[0mm][l]{{\bf #1}}\\}

\newcommand{\cellfirst}{\cellcolor{Red!40}}
\newcommand{\cellsecond}{\cellcolor{Orange!25}}
\newcommand{\cellthird}{\cellcolor{Yellow!25}}

\definecolor{highlightcolor}{HTML}{0071BC}
\newcommand{\naiveedmcolor}[1]{\textcolor{highlightcolor}{#1}}
\newcommand{\xmark}{\ding{55}}%

\newcommand{\netf}{\mathcal{F}_{\theta}}
\newcommand{\netd}{\mathcal{D}_{\theta}}

\newcommand{\noise}{\boldsymbol{\epsilon}} 

\newcommand{\xx}{\boldsymbol{x}} 
\newcommand{\xxh}{\boldsymbol{\hat{x}}} 
\newcommand{\xxt}{\boldsymbol{\tilde{x}}}

\newcommand{\vv}{\boldsymbol{v}}

\newcommand{\upsamplingfactor}{s}
\newcommand{\height}{H}
\newcommand{\width}{W}
\newcommand{\frames}{T}
\newcommand{\patchheight}{H_p}
\newcommand{\patchwidth}{W_p}
\newcommand{\patchframes}{T_p}

\newcommand{\framerate}{\nu}
\newcommand{\resolution}{r}
\newcommand{\cskip}{c_\text{skip}}
\newcommand{\cin}{c_\text{in}}
\newcommand{\cout}{c_\text{out}}
\newcommand{\cnorm}{c_\text{nrm}}

\newcommand{\target}{\mathcal{F}_\text{tgt}}
\newcommand{\lossweight}{\lambda}

\newcommand{\effectivelossweight}{w}
\newcommand{\sigmanoise}{\boldsymbol{\sigma}} 
\newcommand{\sigmadata}{\boldsymbol{\sigma}_\text{data}}
\newcommand{\sigmain}{\boldsymbol{\sigma}_\text{in}}
\newcommand{\sigmamin}{\boldsymbol{\sigma}_\text{min}}
\newcommand{\sigmamax}{\boldsymbol{\sigma}_\text{max}}

\newcommand{\snr}{\mathit{SNR}}

\newcommand{\pdata}{p_\text{data}}
\newcommand{\ptrain}{p_\text{train}}
\newcommand{\boldzero}{\mathbf{0}}
\newcommand{\boldi}{\mathbf{I}}

\begin{document}

\doparttoc 
\faketableofcontents 


\twocolumn[{%
\renewcommand\twocolumn[1][]{#1}%
\maketitle
\begin{center}
    \centering
    \captionsetup{type=figure}
    \includegraphics[width=\textwidth]{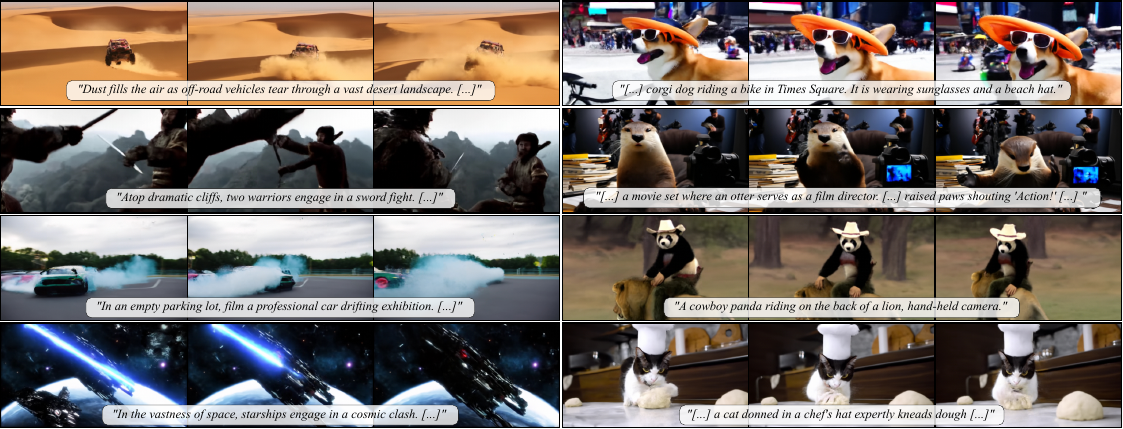}
    \captionof{figure}{Samples produced by the proposed text-to-video generation method for a selection of prompts. Thanks to joint spatiotemporal video modeling, our generator can synthesize temporally coherent videos with large motion (left) while retaining the semantic control capabilities typical of large-scale text-to-video generators (right).  See the \website{} for additional samples.}
\end{center}%
}]

\maketitle

\begin{abstract}

Contemporary models for generating images show remarkable quality and versatility. Swayed by these advantages, the research community repurposes them to generate videos.
Since video content is highly redundant, we argue that naively bringing advances of image models to the video generation domain reduces motion fidelity, visual quality and impairs scalability. In this work, we build \methodname{}, a video-first model that systematically addresses these challenges. To do that, we first extend the EDM framework to take into account spatially and temporally redundant pixels and naturally support video generation. Second,  we show that a U-Net—a workhorse behind image generation—scales poorly when generating videos, requiring significant computational overhead. Hence, we propose a new transformer-based 
architecture that trains 3.31 times faster than U-Nets (and is $\sim$4.5 faster at inference). This allows us to efficiently train a text-to-video model with billions of parameters for the first time, reach state-of-the-art results on a number of benchmarks, and generate videos with substantially higher quality, temporal consistency, and motion complexity. The user studies showed that our model was favored by a large margin over the most recent methods.

\end{abstract}
\makeatletter{\renewcommand*{\@makefnmark}{}
\footnotetext{$^*$
Work performed while interning at Snap Inc.}
\makeatother}

\section{Introduction}
\label{sec:intro}

Creating and sharing visual content is one of the key ways for people to express themselves in the digital world. Accessible to only professionals in the past, the capability to create~\cite{rombach2021highresolution,li2023snapfusion,saharia2022imagen,yu2022scaling} and edit~\cite{ruiz2023dreambooth,brooks2022instructpix2pix,qi2023fatezero} images with stunning quality and realism was unlocked to everyone by the advent of large text-to-image models and their variations. 

Fueled by this progress, large-scale text-to-video models \cite{ho2022imagenvideo,singer2022makeavideo,ge2023preserve,wang2023videofactory,blattmann2023alignyourlatents} are rapidly advancing too.
Current large-scale diffusion-based video generation frameworks are strongly rooted into their image counterparts \cite{blattmann2023alignyourlatents,ge2023preserve}. The availability of consolidated image generation architectures such as U-Nets \cite{ronneberger2015unet} with publicly-available image-pretrained models \cite{rombach2021highresolution} made them a logical foundation onto which to build large-scale video generators with the main architectural modifications focusing on the insertion of ad-hoc layers to capture temporal dependencies \cite{ge2023preserve,blattmann2023alignyourlatents,wang2023videofactory,ho2022imagenvideo,singer2022makeavideo}. Similarly, training is performed under image-based diffusion frameworks with the model being applied both to videos and to a separate set of images to improve the diversity of the results~\cite{ho2022video,singer2022makeavideo,ge2023preserve,ho2022imagenvideo}.

We argue that such an approach is suboptimal under multiple aspects which we systematically address in this work. First, image and video modalities present intrinsic differences given by the similarity of content in successive video frames \cite{ge2023preserve,chen2023importance}. 
By analogy, image and video compression algorithms are based on vastly different approaches \cite{ma2019imageav}.
To address this issue, we rewrite the EDM \cite{karras2022edm} framework with a focus on high-resolution videos. 
Differently from past work where videos were treated as a sequence of images, we perform joint video-image training by treating images as \emph{high frame-rate videos} to avoid modality mismatches introduced by the absence of the temporal dimension within purely image-based training.
Second, the widely adopted U-Net \cite{ronneberger2015unet} architecture is required to fully processes each video frame. This increases computational overhead compared to purely text-to-image models, posing a very practical limit on model scalability. The latter is a critical factor in obtaining high-quality of results~\cite{ho2022imagenvideo,ge2023preserve}. Extending U-Net-based architectures to naturally support spatial and temporal dimensions requires volumetric attention operations, which have prohibitive computational demands. Inability to do so affects the outputs, resulting in \emph{dynamic images} or motion artifacts being generated instead of videos with coherent and diverse actions. 

Following our compression analogy, we propose to leverage repetition between frames and introduce a scalable transformer architecture that treats spatial and temporal dimensions as a single, compressed, 1D latent vector. This highly compressed representation allows us to perform spatio-temporal computation jointly and enables modelling of complex motions. Our architecture is inspired by FIT~\cite{chen2023fit}, which we scale to billions of parameters for the first time. Compared to U-Nets, our model features a significant $3.31 \times$ reduction in training time and $4.49 \times$ reduction in inference time while achieving higher generation quality.

We evaluate \methodname{} on the widely-adopted UCF101 \cite{soomro2012ucf} and MSR-VTT \cite{xu2016msrvtt} datasets.
Our generator shows state-of-the-art performance across the range of benchmarks with particular regard to the quality of the generated motion. Most interestingly, we performed a number of user studies against the most recent open- and close-source methods and found that according to the participants of the study our model features photorealism comparable to Gen-2 \cite{esser2023structure}, while being significantly better than Pika \cite{pika} and Floor33 \cite{he2023latent}. Most excitedly, the preference of user-study participants favoured \methodname{} by a large margin when text alignment and motion quality were assessed. Compared to Gen-2 \cite{esser2023structure} on prompt-video alignment our model was preferred in 81\% of cases (80\% against Pika \cite{pika}, 81\% against Floor33 \cite{he2023latent}), generated most dynamic videos with most amount of motion (96\% against Gen2 \cite{esser2023structure}, 89\% against Pika \cite{pika}, 88\% against Floor33 \cite{he2023latent}) and had the best motion quality (79\% against Gen-2 \cite{esser2023structure}, 71\% against Pika \cite{pika}, 79\% against Floor33 \cite{he2023latent}).

\section{Related Work}
\label{sec:related_work}


\noindent\textbf{Video Generation}~Video generation is a challenging and long-studied task. Due to its complexity, a large number of works focus on modeling narrow domains \cite{tian2021good,tulyakov2018moco,sihyun2022digan,skorokhodov2022styleganv,saito2020tganv2,lee2018savp,clark2019dvdgan,shen2023mostganv,brooks2022generating,yan2021videogpt,srivastava2015unsupervised,moing2021ccvs,ge2022long,yu2023magvit} and adopt adversarial training \cite{tian2021good,tulyakov2018moco,sihyun2022digan,skorokhodov2022styleganv,saito2020tganv2,lee2018savp,clark2019dvdgan,shen2023mostganv,brooks2022generating} or autoregressive generation techniques \cite{yan2021videogpt,srivastava2015unsupervised,moing2021ccvs,ge2022long,yu2023magvit}. 
To address the narrow domain limitation, the task of text-to-video generation was proposed \cite{mittal2017syncdraw} and both autoregressive models \cite{mittal2017syncdraw,hong2022cogvideo,wu2021godiva,chenfei2022nuwa,villegas2022phenaki} and GANs \cite{li2018videogeneration} emerged. 

The recent success of diffusion models in the context of text-to-image generation \cite{saharia2022imagen,rombach2021highresolution,balaji2022ediffi} fostered tremendous progress in the task \cite{ho2022video,yin2023nuwaxl,ho2022imagenvideo,singer2022makeavideo,ge2023preserve,wang2023videofactory,he2023latent,zhou2023magicvideo,blattmann2023alignyourlatents,luo2023videofusion,an2023latentshift,guo2023animatediff}. ImagenVideo \cite{ho2022imagenvideo} and Make-A-Video \cite{singer2022makeavideo} propose a deep cascade of temporal and spatial upsamplers to generate videos and jointly train their models on image and video datasets. PYoCo \cite{ge2023preserve} introduces a correlated noise model to capture similarities between video frames. Video LDM \cite{blattmann2023alignyourlatents} adopts a latent diffusion paradigm where a pre-trained latent image generator and latent decoder are finetuned to generate temporally coherent videos. AnimateDiff \cite{guo2023animatediff} freezes a pre-trained latent image generator and trains only a newly inserted motion modeling module. These works employ U-Nets with separable spatial and temporal computation which poses a limitation on motion modeling capabilities. VideoFactory \cite{wang2023videofactory} improves upon this paradigm by proposing a Swapped Spatiotemporal Cross-Attention that improves interactions between the spatial and temporal modalities along 3D windows.

Differently from this corpus of works which adapts the U-Net \cite{ronneberger2015unet} architecture to the video generation task, we show that employing transformer-based FIT \cite{chen2023fit} architectures results in significant training time savings, scalability improvements, and performance increase thanks to their learnable compressed video representation. In particular, we show that the global joint spatiotemporal modeling strategy enabled by our compressed video representation results in significant improvements in temporal consistency and motion modeling capabilities. \\

\noindent\textbf{High-Resolution Generation}~Different approaches have been proposed to enable the generation of high-resolution outputs.
Cascaded diffusion models \cite{ho2022imagenvideo,saharia2022imagen,balaji2022ediffi,ge2023preserve,singer2022makeavideo} adopt a set of independent diffusion models designed to successively upsample the results of the previous step. 
Latent diffusion models \cite{rombach2021highresolution,he2023latent,blattmann2023alignyourlatents,zhou2023magicvideo,an2023latentshift} make use of a pretrained autoencoder to encode the input into a low-dimensional set of latent vectors and learn a diffusion model on this latent representation.

A different family of methods generates high-resolution outputs end-to-end without employing cascades of models or latent diffusion. Simple Diffusion \cite{hoogeboom2023simplediffusion} and \emph{Chen} \cite{chen2023importance} directly generate high-resolution images by adapting the noise schedule of the diffusion process.
f-DM \cite{gu2022fdm} and RDM \cite{teng2023relay} design a diffusion process that seamlessly transitions between different resolutions. MDM \cite{gu2023matryoshka} proposes a strategy where a single model is trained to simultaneously denoise inputs at progressively higher resolutions.

In this work, we adopt a two-stage cascaded model out of two considerations: (i) it avoids temporal inconsistencies in the forms of flickering of high-frequency details that may be introduced by latent autoencoders \cite{blattmann2023alignyourlatents}, (ii) it increases model capacity with respect to an end-to-end model by creating two specialized models, one for the low resolution focusing on motion modeling and scene structure, and one for the high-resolution, focusing on high-frequency details.\\  


\noindent\textbf{Diffusion Frameworks}~Diffusion generative models are a set of techniques modeling generation as a pair of processes: a forward process progressively destructing a sample with noise, and a reverse process modeling generation as the progressive denoising of a sample. Different formulations of diffusion models have been proposed in the literature. Denoising Diffusion Probabilistic Models (DDPMs) \cite{ho2020ddpm,sohl2015deepunsupervised} formulate the forward and backward process as Markov chains. Score-based Generative Models (SGMs) \cite{song2019generative,song2020improved} model the score of the probability density function of a series of data distributions perturbed with increasing levels of noise, \ie the direction of largest increase in the data log probability density function. An avenue of works~\cite{song2021maximum,song2021scorebased} generalizes DDPMs and SGMs to infinite noise levels through Stochastic Differential Equations (SDEs).
In this work, we adopt the SGM framework of EDM \cite{karras2022edm} which we reformulate for the generation of high-resolution videos.
 
\section{Method}
\label{sec:method}

We propose the generation of high-resolution videos by rewriting the EDM \cite{karras2022edm} diffusion framework for high-dimensional inputs and proposing an efficient transformer architecture based on FITs \cite{chen2023fit} which we scale to billions of parameters and tens of thousands input patches. Sec.~\ref{sec:edm_introduction} provides an introduction to the EDM framework, Sec~\ref{sec:edm_for_high_resolution_video_generation} highlights the challenges of applying diffusion frameworks to high dimensional inputs and proposes a revisited EDM-based diffusion framework. Sec.~\ref{sec:image_video_modality_matching} proposes a method to reduce the gap between image and video modalities for joint training. Finally, Sec.~\ref{sec:architecture} describes our scalable video generation architecture, while Sec.~\ref{sec:training} and Sec.~\ref{sec:inference} respectively describe the training and inference procedures.

\subsection{Introduction to EDM}
\label{sec:edm_introduction}
Diffusion models have achieved remarkable success in image and video generation. Among the proposed frameworks, \emph{Karras} \etal{} \cite{karras2022edm} provide a unified view of common diffusion frameworks and formulate EDM.
EDM defines a variance-exploding forward diffusion process $p(\xx_{\sigmanoise}|\xx) \sim \mathcal{N}(\xx, \sigmanoise^2 \boldi)$, where $\sigmanoise \in [\sigmamin, \sigmamax]$ represents the diffusion timestep coinciding with the standard deviation of the applied noise, and $\xx_{\sigmanoise}$ represents the data at the current noise level. A denoiser function $\netd$ is learned to model the reverse process using the denoising objective:
\vspace{-1mm}
\begin{equation}
  \mathcal{L}(\netd) = \mathbb{E}_{\sigmanoise, \xx, \noise} \Big[ \lossweight(\sigmanoise) ~\big\lVert \netd(\xx_{\sigmanoise}) - \xx \big\rVert^2_2 \Big]
  \label{eq:netd_loss}
  \text{,}
  \vspace{-1mm}
\end{equation}
where $\lossweight$ is the loss weighting function, $\xx \sim \pdata$ is a data sample, $\noise$ is gaussian noise, and $\sigmanoise \sim \ptrain$ is sampled from a training distribution. $\netd(\xx_{\sigmanoise})$ is defined as:
\vspace{-1mm}
\begin{equation}
    \netd(\xx_{\sigmanoise}) = \cout(\sigmanoise) \netf\left(\cin(\sigmanoise) \xx_{\sigmanoise}\right) + \cskip(\sigmanoise)\xx_{\sigmanoise}
    \label{eq:netd}
    \text{,}
\vspace{-1mm}
\end{equation}
where $\netf$ is a neural network, and $\cout$, $\cskip$ and $\cin$ represent scaling functions.
In particular, the denoising objective $\mathcal{L}(\netf)$ can equivalently be expressed in terms of $\netf$ as: 
\vspace{-1mm}
\begin{equation}
\label{eq:netf_loss}
\resizebox{75mm}{!}{
$\mathcal{L}(\netf) = \mathbb{E}_{\sigmanoise, \xx, \noise} \Big[ \effectivelossweight(\sigmanoise) ~\big\lVert \netf(\cin(\sigmanoise) \xx_{\sigmanoise}) - \cnorm(\sigmanoise) \target \big\rVert^2_2 \Big]
\text{,}$
}
\vspace{-1mm}
\end{equation}
where $\target$ represents the training target, $\cnorm$ is a normalization factor, and $\effectivelossweight$ is a weighting function. These forms, derived in \apref{ap:edm_derivation}, are presented in Tab.~\ref{table:diffusion_process_comparison}. 

A second order Runge-Kutta sampler is proposed to reverse the diffusion process and produce sample $\xx$ starting from gaussian noise $\xx_{\sigmamax} \sim \mathcal{N}(\boldzero, \sigmamax^2 \boldi)$.

\begin{table}
\begin{center}

\setlength{\tabcolsep}{0.0pt}
\footnotesize
\resizebox{\columnwidth}{!}{%
\begin{tabular}{l@{\hspace{-1mm}}r@{\hspace{3mm}}c@{\hspace{-4mm}}c}
\toprule
 & &  EDM \cite{karras2022edm} & Our \\
\midrule
\vspace*{.1ex}\titlerowww{Training and Losses}
Forw. process & $\xx_{\sigmanoise}$ & $\xx / \naiveedmcolor{\sigmain} + \sigmanoise \noise$ & $\xx / \naiveedmcolor{\sigmain} + \sigmanoise \noise$ \\
Training target & $\target$ & $\sigmanoise \xx - \sigmadata^2 \noise + \naiveedmcolor{\frac{\sigmadata^2 (\sigmain - 1)}{\sigmain \sigmanoise} \xx}$ & - $\sigmanoise \xx + \sigmadata^2 \noise$ \\
\vspace*{1mm}Eff. loss weigh. & $\effectivelossweight(\sigmanoise)$ & $1$ & $\naiveedmcolor{(\sigmanoise^2 + \sigmadata^2)^2 / (\sigmanoise^2 + \frac{\sigmadata^2}{\sigmain})^2}$ \\
Loss weigh. & $\lossweight(\sigmanoise)$ & $1 / \sigmadata^2 + 1 / \sigmanoise^2$ & $1 / \sigmadata^2 + 1 / \sigmanoise^2$ \\

\midrule
\vspace*{.1ex}\titlerowww{Network Parametrization}
\vspace*{1mm}Input scaling & $\cin(\sigmanoise)$ & $1 / \sqrt{\smash[b]{\sigmadata^2 + \sigmanoise^2}}$ & $1 / \sqrt{\smash[b]{\sigmadata^2 / \naiveedmcolor{\sigmain^2} + \sigmanoise^2}}$ \\
\vspace*{1mm}Output scaling & $\cout(\sigmanoise)$ & $\frac{\sigmanoise \sigmadata^2}{\sqrt{\smash[b]{\sigmanoise^2 + \sigmadata^2}}}$ & $-\naiveedmcolor{\sigmain}\sigmanoise \sigmadata \frac{ \sqrt{\smash[b]{\sigmanoise^2 + \sigmadata^2}}}{\sigmadata^2 + \naiveedmcolor{\sigmain}\sigmanoise^2}$ \\
\vspace*{1mm}Skip scaling & $\cskip(\sigmanoise)$ & $\frac{\sigmadata^2}{\sigmanoise^2 + \sigmadata^2}$ & $\frac{\naiveedmcolor{\sigmain}\sigmadata^2}{\naiveedmcolor{\sigmain}\sigmanoise^2 + \sigmadata^2}$ \\
Target scaling & $\cnorm(\sigmanoise)$ & $1 / \sigmadata \sqrt{\smash[b]{\sigmanoise^2 + \sigmadata^2}}$ & $1 / \sigmadata \sqrt{\smash[b]{\sigmanoise^2 + \sigmadata^2}}$ \\

\bottomrule
\end{tabular}
}
\end{center}
\caption{Definitions of functions in Eq.~\eqref{eq:netd_loss}, Eq.~\eqref{eq:netd} and Eq.~\eqref{eq:netf_loss} for the EDM and our proposed diffusion framework as derived in \apref{ap:edm_derivation} and \apref{ap:our_framework_derivation}, where we \naiveedmcolor{highlight} the terms induced by the input scaling factor $\sigmain$. Our framework is equivalent to EDM for $\sigmain=1$ but avoids the unstable term $\frac{\sigmadata^2 (\sigmain - 1)}{\sigmain \sigmanoise} \xx$ induced by $\sigmain\neq1$ in $\target$. This form highlights that the train target and loss weight match the $\vv$-prediction \cite{salimans2022progressive} framework for $\sigmadata=1$. All other framework parameters are unaltered with respect to EDM.} 
\label{table:diffusion_process_comparison}

\end{table}

\subsection{EDM for High-Resolution Video Generation}
\label{sec:edm_for_high_resolution_video_generation}

\begin{figure}
\includegraphics[width=\columnwidth]{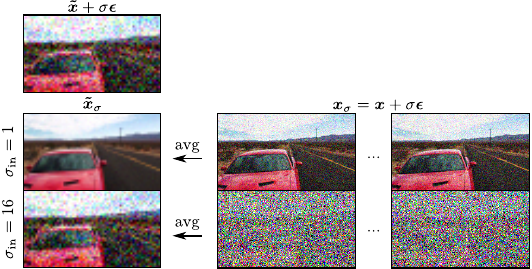}
  \caption{
  \textbf{Analysis of Signal-to-Noise Ratio ($\snr$)}. Top: noise $\sigmanoise$ is applied to an image. Middle: the same noise $\sigmanoise$ is applied to a 16-frames-long video $\xx$ without scaling. A clean image can be easily restored by simply taking average, indicating an increased $\snr$. Bottom: to maintain the original $\snr$, we scale down the 16 frames by $\sigmain$ before noise application. Averaging is not able to restore the images, indicating the $\snr$ is maintained as $\xxt + \sigma \noise$.
  }
  \label{fig:noise}
\end{figure}

EDM is originally proposed as an image generation framework and its parameters are optimized for $64 \times 64$px image generation. 
Alterations in spatial resolution or the introduction of videos with shared content between frames allow the denoising network to trivially recover a noisy frame in the original resolution with higher signal-to-noise-ratio ($\snr$), which the original framework was designed to see at lower noise levels. To see why, consider a noisy video $\xx_{\sigmanoise} \in \mathbb{R}^{\frames \times \upsamplingfactor \cdot \height \times \upsamplingfactor \cdot \width} \sim \mathcal{N}(\xx, \sigmanoise^2 \boldi)$ 
where $\frames$ is the number of frames and $\upsamplingfactor$ is an upsampling factor. We build the corresponding clean and noisy frames at original resolution $\xxt, \xxt_{\sigmanoise} \in \mathbb{R}^{1 \times \height \times \width}$ by averaging values in each $\frames \times \upsamplingfactor \times \upsamplingfactor$ block of pixels. 
As a consequence of averaging, the noise variance is reduced by a factor $\frames \upsamplingfactor^2$, \ie $\xxt_{\sigmanoise} \sim \mathcal{N}(\xxt, \frac{\sigmanoise^2}{\frames \upsamplingfactor^2} \boldi)$, thus $\xxt_{\sigmanoise}$ has an increased signal-to-noise-ratio with respect to $\xx_{\sigmanoise}$ (see Fig.~\ref{fig:noise}): $SNR_{\xxt_{\sigmanoise}}=\frames \upsamplingfactor^2 SNR_{\xx_{\sigmanoise}}$. If  pixels in each block share similar content, a typical situation in high-resolution videos, then the information in the averaged frame is useful for recovering $\xx$ and can be exploited at training time by the denoiser function. This creates a train-inference mismatch during the initial sampling steps as the average frame does not yet contain a well-formed signal, yet the denoiser is reliant on its presence. Thus, for best performance, any alteration to $\frames$ or $\upsamplingfactor$ should instead maintain the same signal-to-noise ratio at the original resolution for which the diffusion framework was designed.

To restore the optimal $\snr$ at the original resolution, the magnitude of the input signal can be reduced \cite{chen2023importance} by a corresponding factor $\sigmain=\upsamplingfactor \sqrt{\frames}$ as illustrated in Fig.~\ref{fig:noise}. Consequently, we redefine the forward process as $p(\xx_{\sigmanoise}|\xx) \sim \mathcal{N}(\xx / \sigmain, \sigmanoise^2 \boldi)$. 

We rewrite the EDM framework to introduce the input scaling factor in \apref{ap:our_framework_derivation} and highlight the changes in Tab.~\ref{table:diffusion_process_comparison}. We notice that a naive introduction of the scaling factor would alter the training target $\target$ in a way that makes the objective explode for small noise values (see \apref{ap:edm_derivation}). We thus leverage the training objective expressed in the form of Eq.~\eqref{eq:netf_loss} to rewrite the EDM process in a way that ensures $\target$ remains unchanged, the effective loss weight $\effectivelossweight(\sigmanoise)$ is such that it keeps the loss weight $\lossweight(\sigmanoise)$ unchanged, $\cin(\sigmanoise)$ and $\cnorm(\sigmanoise)$ normalize the input and training target to have unit variance, and the framework is equivalent to the original EDM formulation for $\sigmain=1$ (see \apref{ap:our_framework_derivation}).

Finally, we modify the sampler according to the newly defined forward process that requires the signal component in $\xx_{\sigmanoise}$ to be scaled by $\sigmain$. This is achieved by dividing the $\netd(\xx_{\sigmanoise})$ by $\sigmain$ and multiplying the final denoised sample $\xx_0$ by $\sigmain$ to restore the signal magnitude.

\begin{figure*}
    \begin{subfigure}{0.43\textwidth}
        \includegraphics[width=\textwidth]{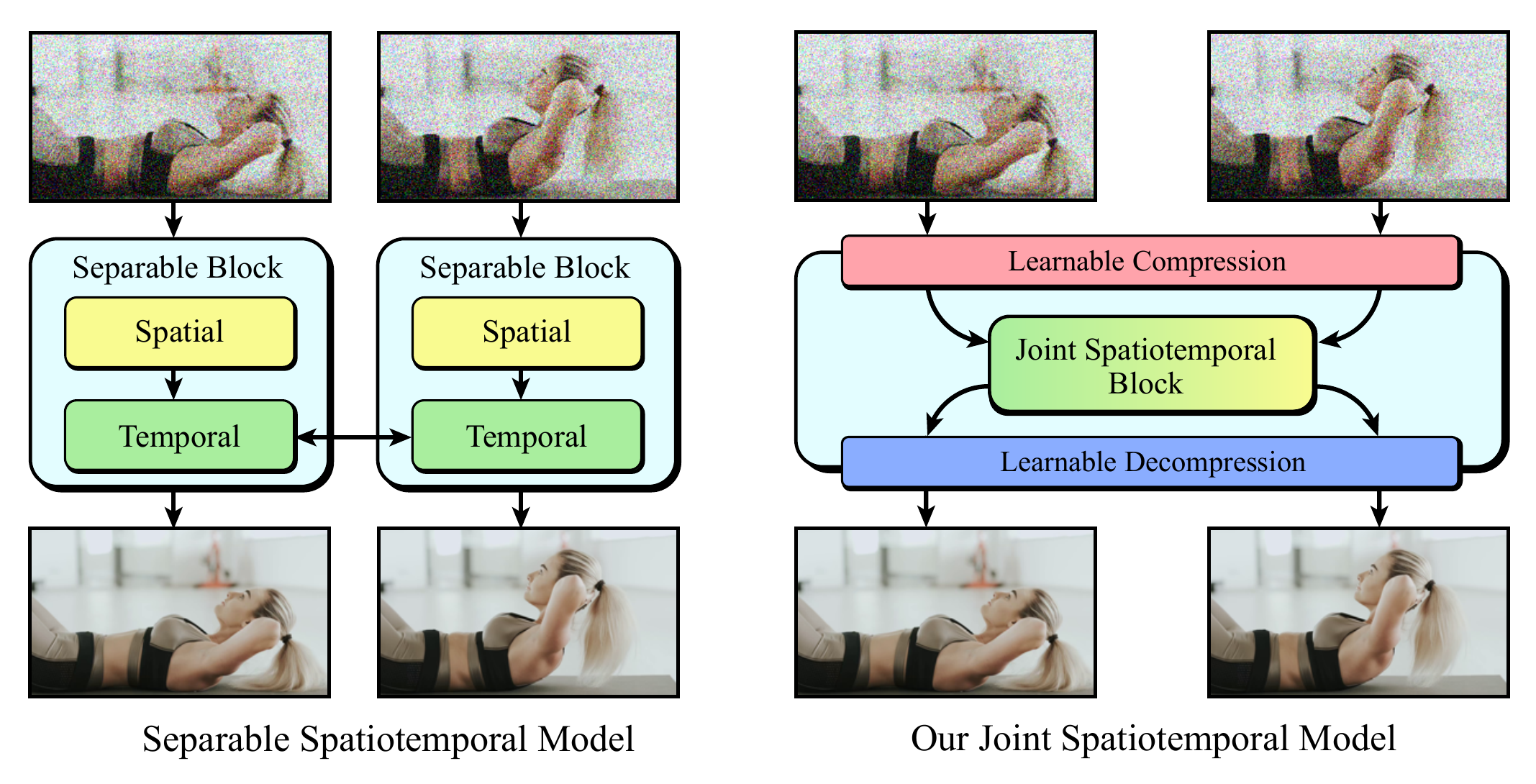}
        \caption{Computational Paradigms for Videos}
        \label{fig:spatiotemporal_computation}
    \end{subfigure}
    \hfill
    \begin{subfigure}{0.53\textwidth}
        \includegraphics[width=\textwidth]{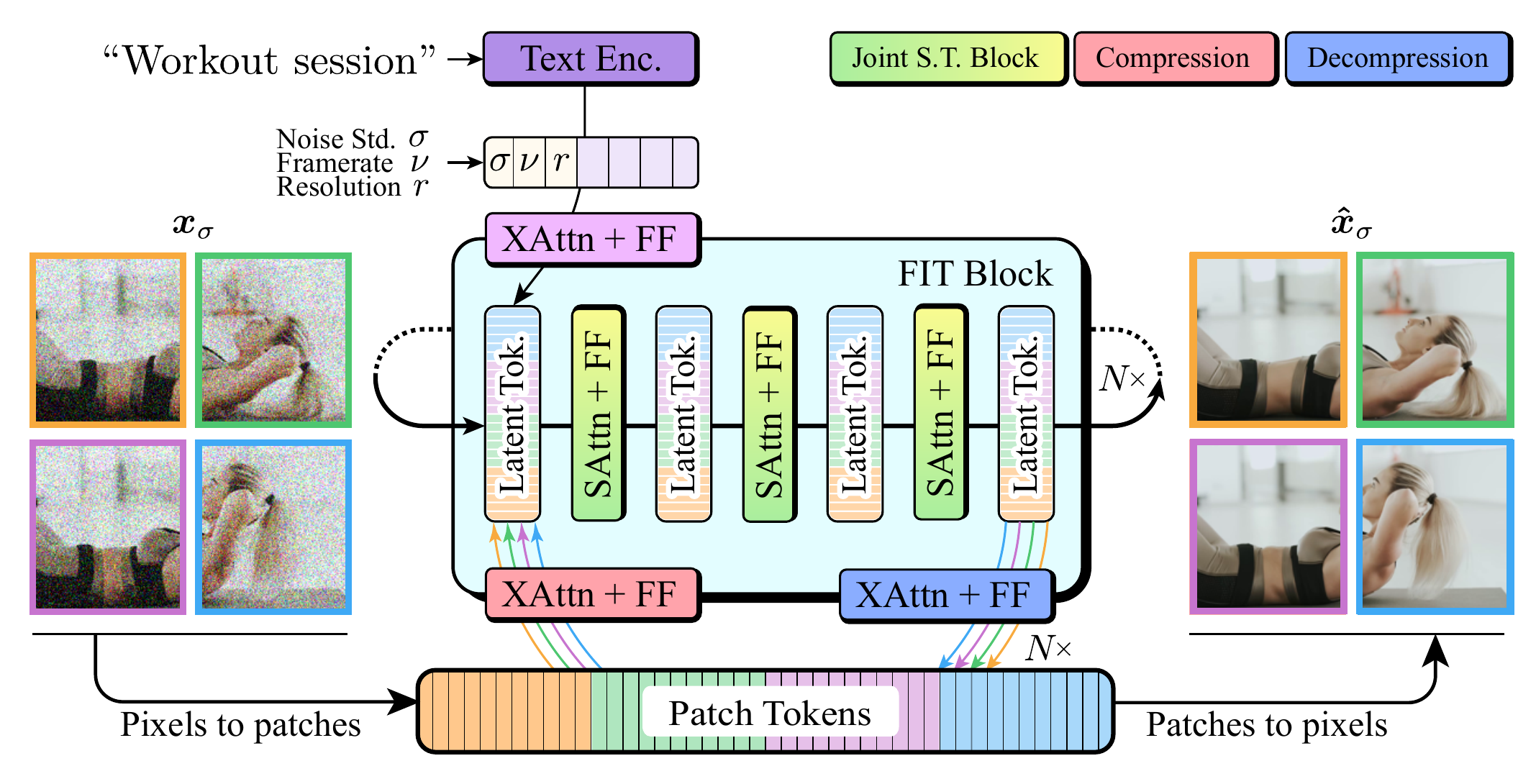}
        \caption{\methodname{} FIT Architecture}
        \label{fig:architecture}
    \end{subfigure}
  \caption{(a-left) U-Net-based text-to-image architectures are adapted to do video generation by inserting temporal layers applied sequentially with spatial layers, creating separable spatiotemporal blocks. Spatial computation is repeated for each frame independently, limiting scalability. (a-right) Our scalable transformer-based model jointly performs spatial and temporal computation on a learnable compressed video representation for improved motion modeling and scalability. (b) The proposed \methodname{} FIT architecture. Given a noisy input video $\xx_{\sigmanoise}$, the model estimates the denoised video $\xxh_{\sigmanoise}$ by recurrent application of FIT blocks. Each block reads information from the patch tokens into a small set of latent tokens on which computation is performed. The results are written to the patch tokens. Conditioning information in the form of text embeddings, noise level $\sigma$, frame-rate $\framerate$ and resolution $\resolution$ is provided through an additional read operation.} 
  \label{fig:architecture_and_computation}
\end{figure*}


\subsection{Image-Video Modality Matching}
\label{sec:image_video_modality_matching}
Due to the limited amount of captioned video data with respect to images, joint image-video training is widely adopted \cite{ho2022video,ge2023preserve,ho2022imagenvideo,singer2022makeavideo} with the same diffusion process typically applied to both modalities. However, as shown in Sec.~\ref{sec:edm_for_high_resolution_video_generation}, the presence of $\frames$ frames in videos calls for a different process with respect to an image with the same resolution. A possibility would be to adopt different input scaling factors for the two modalities. We argue that this solution is undesirable in that it increases the complexity of the framework and image training would not foster the denoising model to learn temporal reasoning, a fundamental capability of a video generator. To sidestep these issues while using a unified diffusion process, we match the image and video modalities by treating images as $\frames$ frames videos with infinite frame-rate and introduce a variable frame-rate training procedure blending the gap between the image and video modalities.

\subsection{Scalable Video Generator}
\label{sec:architecture}

U-Nets \cite{ronneberger2015unet} have shown success in video generation where they are typically augmented with temporal attention or convolutions for modeling the temporal dimension \cite{ge2023preserve,ho2022imagenvideo,singer2022makeavideo,blattmann2023alignyourlatents,ho2022video}.
However, such an approach requires a full U-Net forward pass for each of the $\frames$ video frames, rapidly becoming prohibitively expensive (see Fig.~\ref{fig:spatiotemporal_computation}). These factors pose a practical limit on model scalability---a primary factor in achieving high generation quality \cite{singer2022makeavideo,ho2022imagenvideo,ge2023preserve,he2023latent}---and similarly limit possibilities for joint spatio-temporal modeling \cite{wang2023videofactory}. We argue that treating spatial and temporal modeling in a separable way \cite{ho2022imagenvideo,singer2022makeavideo,ge2023preserve,blattmann2023alignyourlatents} causes motion artifacts, temporal inconsistencies or generation of \emph{dynamic images} rather than videos with vivid motion. 
Video frames, however, contain spatially and temporally redundant content that is amenable to compression \cite{ma2019imageav}. We argue that learning and operating on a compressed video representation and jointly modeling the spatial and temporal dimensions are necessary steps to achieve the scalability and motion-modeling capabilities required for high-quality video generation.

FITs \cite{chen2023fit} are efficient transformer-based architectures that have recently been proposed for high-resolution image synthesis and video generation.  Their main idea, summarized in Fig.~\ref{fig:architecture_and_computation} is that of learning a compressed representation of their input through a set of learnable latent tokens and of focusing computation on this learnable latent space, allowing input dimensionality to grow with little performance penalty. First, FITs perform patchification of the input and produce a sequence of patch tokens which are later divided into groups. A set of latent tokens is then instantiated and a sequence of computational blocks is applied. Each block first performs a cross attention ``read'' operation between latent tokens and conditioning signals such as the diffusion timestep, then an additional groupwise ``read'' cross attention operation between latent and patch tokens of corresponding groups to compress patch information, applies a series of self attention operations to the latent tokens, and performs a groupwise ``write'' cross attention operation that decompresses information in the latent tokens to update the patch tokens. Finally, the patch tokens are projected back to the pixel space to form the output. Self conditioning is applied on the set of latent tokens to preserve the compressed video representation computed in previous sampling steps.

While promising, these architectures have not yet been scaled to the billion-parameters size of state-of-the-art U-Net-based video generators, nor they have been applied to high-resolution video generation. In the following, we highlight the architectural considerations necessary to achieve these goals. Temporal modeling is a fundamental aspect of a high-quality video generator. FITs produce patch tokens by considering three dimensional patches of size $\patchframes \times \patchheight \times \patchwidth$ spanning both the spatial and temporal dimensions. We find values of $\patchframes > 1$ to limit temporal modeling performance, so we consider patches spanning the spatial dimension only. In addition, similarly to patches, FITs group patch tokens into groups spanning both the temporal and spatial dimensions, and perform cross attention operations group by group. We observe that the temporal size of each group should be configured so that each group covers all $\frames$ video frames for best temporal modeling. Furthermore, videos contain more information with respect to images due to the presence of the temporal dimension, thus we increase the number of latent tokens representing the size of the compressed space in which joint spatiotemporal computation is performed. Finally, FITs make use of local layers which perform self attention operations on patch tokens corresponding to the same group. We find this operation to be computationally expensive for large amounts of patch tokens (147.456 for our largest resolution) and replace it with a feed forward module after each cross attention ``read'' or ``write'' operation. 

Our model makes use of conditioning information represented by a sequence of conditioning tokens to control the generation process. In addition to the token representing the current $\sigmanoise$, to enable text conditioning, we introduce a T5-11B \cite{raffel2022exploring} text encoder extracting text embeddings from the input text. To support variable video framerates and large differences in resolution and aspect ratios in the training data, we concatenate additional tokens representing the framerate and original resolution of the current input.

To generate high-resolution outputs, we implement a model cascade consisting of a first-stage model producing $36 \times 64$px videos and a second-stage upsampling model producing $288 \times 512$px videos. To improve upsampling quality, we corrupt the second-stage low-resolution inputs with a variable level of noise during training \cite{ho2022imagenvideo,saharia2022imagen} and during inference apply a level of noise to the first-stage outputs obtained by hyperparameter search. 

We present detailed model hyperparameters in \apref{ap:architecture_details}.

\subsection{Training}
\label{sec:training}

We train \methodname{} using the LAMB \cite{you2020large} optimizer with a learning rate of $5e^{-3}$, a cosine learning schedule and a total batch size of 2048 videos and 2048 images, achievable thanks to our scalable video generator architecture. We train the first-stage model over 550k steps and finetune the second-stage model on high-resolution videos starting from the first-stage model weights for 370k iterations.
Following the observations in Sec~\ref{sec:edm_for_high_resolution_video_generation}, we pose $\sigmain=\upsamplingfactor\sqrt{\frames}$. Considering videos with $\frames=16$ frames and the original $64$px resolution for which EDM was designed, we set $\sigmain=4$ for the first-stage and $\sigmain=32$ for the second-stage model.

We present training details and parameters in \apref{ap:training_details}.

\subsection{Inference}
\label{sec:inference}

We produce video samples from gaussian noise and user-provided conditioning information using the deterministic sampler of \cite{karras2022edm} and our two-stage cascade. We use 256 sampling steps for the first-stage and 40 for the second-stage model, and employ classifier free guidance \cite{ho2022classifierfree} to improve text-video alignment (see \apref{ap:additional_ablations}) unless otherwise specified. We find dynamic thresholding \cite{saharia2022imagen} and oscillating guidance \cite{ho2022imagenvideo} to consistently improve sample quality.

\section{Evaluation}
\label{sec:evaluation}
In this section, we perform evaluation of \methodname{} against baselines and validate our design choices. Sec.~\ref{sec:datasets} introduces the employed datasets, Sec.~\ref{sec:evaluation_protocol} defines the evaluation protocol, Sec.~\ref{sec:ablations} shows ablations of our diffusion framework and architectural choices, Sec.~\ref{sec:comparison_to_baselines} quantitatively compares our method to state-of-the-art large-scale video generators and Sec.~\ref{sec:qualitative_evaluation} performs qualitative evaluation. We complement evaluation by showcasing samples in the \supp{} and \website{}.

\subsection{Datasets}
\label{sec:datasets}
We train our models on an internal dataset consisting of 1.265M images and 238k hours of videos, each with a corresponding text caption.
Due to the difficulty in acquiring high-quality captions for videos, we develop a video captioning model that we use to produce synthetic video captions for the portion of videos in the dataset missing such annotation.

We make use of the following datasets for evaluation which are never observed during training:

\noindent\textbf{UCF-101} \cite{soomro2012ucf} is a video dataset containing 13.320 $320\times240$px Youtube videos from 101 action categories.\\
\noindent\textbf{MSR-VTT} \cite{xu2016msrvtt} is a dataset containing 10.000 $320\times240$px web-crawled videos, each manually annotated with 20 text captions. The test set contains 2.990 videos and 59.800 corresponding captions.\\

\begin{table}
\begin{center}

\setlength{\tabcolsep}{3.0pt}
\footnotesize
\resizebox{\columnwidth}{!}{%
\begin{tabular}{lcccccc}
\toprule
 &  FID $\downarrow$ & FVD $\downarrow$ & CLIPSIM $\uparrow$ & Train Thr. $\downarrow$ & Inf. Thr. $\downarrow$ \\
\midrule
 U-Net 85M \cite{dhariwal2021diffusion} & 8.21 & 45.94 & 0.2319 & \cellsecond{}133.2 & \cellsecond{}49.6 \\
 U-Net 284M \cite{dhariwal2021diffusion} & \cellthird{}4.90 & \cellsecond{}23.76 & \cellthird{}0.2391 & \cellthird{}230.3 & \cellthird{}105.1 \\
 \methodname{} FIT 500M & \cellsecond{}3.07 & \cellthird{}27.79 & \cellsecond{}0.2459 & \cellfirst{}69.5 & \cellfirst{}23.4 \\
 \methodname{} FIT 3.9B & \cellfirst{}2.51 & \cellfirst{}12.31 & \cellfirst{}0.2579 & 526.0 & 130.4 \\
\bottomrule
\end{tabular}
}
\end{center}
\caption{Performance of different architectures and model sizes on our internal dataset in $64 \times 36$px resolution. We observe strong performance gains with scaling and note that FITs present better performance with improved speed with respect to U-Nets. Train and inference throughputs in ms/video/GPU.}
\label{table:ablation_architecture_scaling}

\end{table}

\begin{table}
\begin{center}

\setlength{\tabcolsep}{3.0pt}
\footnotesize
\begin{tabular}{lcccccc}
\toprule
 & $\sigmadata$ & $\sigmain$ & Imgs. as Videos & FID $\downarrow$ & FVD $\downarrow$ & CLIPSIM $\uparrow$ \\
\midrule
 (i) & 0.5 & 1.0 & $\checkmark$ & 6.58 & 39.95 & 0.2370 \\
 (ii) & 0.5 & 4.0 & $\checkmark$ & 4.03 & 31.00 & 0.2449 \\
   \midrule
 (iv) & 1.0 & 2.0 & $\checkmark$ & 4.45 & 34.89 & 0.2428 \\
  \midrule
 (iii) & 1.0 & 1/4.0 & \xmark & 3.50 & 24.88 & 0.2469 \\
  \midrule
 Ours & 1.0 & 4.0 & $\checkmark$ &  3.07 & 27.79 & 0.2459 \\
\bottomrule
\end{tabular}
\end{center}
\caption{Ablation of different diffusion process configurations varying $\sigmadata$, input scaling $\sigmain$, and treatment of images as infinite-framerate videos, evaluated on our internal dataset in $64 \times 36$px resolution.}
\label{table:ablation_diffusion}

\end{table}

\subsection{Evaluation Protocol}
\label{sec:evaluation_protocol}
To validate the choices operated on the diffusion framework and on model architecture, present method ablations performed in $64 \times 36$px resolution using the first-stage model only, and compute FID \cite{heusel2017advances}, FVD \cite{unterthiner2018towards} and CLIPSIM \cite{wu2021godiva} metrics against the test set of our internal dataset on 50k generated videos.

To evaluate our method against baselines, we follow the protocols highlighted in \cite{singer2022makeavideo,ge2023preserve,wang2023videofactory,blattmann2023alignyourlatents,zhou2023magicvideo,luo2023videofusion} for zero-shot evaluation on the UCF-101 \cite{soomro2012ucf} and MSR-VTT \cite{xu2016msrvtt} datasets. We generate 16 frames videos in $512 \times 288$px resolution at 24fps for all settings. We evaluate both at the native $512 \times 288$px resolution with 16:9 aspect ratio and in the $288 \times 288$px square aspect ratio typically employed on these benchmarks. 
We note that the evaluation protocols of \cite{singer2022makeavideo,ge2023preserve,wang2023videofactory,blattmann2023alignyourlatents,zhou2023magicvideo,luo2023videofusion} present different choices regarding the number of generated samples, distribution of class labels, choice of text prompts. We make use of the following evaluation parameters:

\noindent\textbf{Zero-shot UCF-101} \cite{soomro2012ucf} We generate 10.000 videos \cite{wang2023videofactory,blattmann2023alignyourlatents} sampling classes with the same distribution as the original dataset. We produce a text prompt for each class label \cite{ge2023preserve} and compute FVD \cite{unterthiner2018towards} and Inception Score \cite{salimans2016improved}.\\
\noindent\textbf{Zero-shot MSR-VTT} \cite{xu2016msrvtt} We generate a video sample for each of the 59.800 test prompts \cite{singer2022makeavideo,ge2023preserve} and compute CLIP-FID \cite{kynk2023theroleofimagenet} and CLIPSIM \cite{wu2021godiva}.\\

To provide a more complete performance assessment and compare against state-of-the-art closed-source methods not reporting results for these benchmarks, we perform a user study evaluating photorealism, video-text-alignment and, most importantly, the quantity and quality of the generated motion, important characteristics of a video generator that may signal the generation of \emph{dynamic images}, \ie videos with dim motion, or motion artifacts rather than videos with vivid and high-quality motion.

\subsection{Ablations}
\label{sec:ablations}

To evaluate the proposed FIT architecture, we consider the U-Net of \cite{dhariwal2021diffusion}, which we adapt to the video generation setting by interleaving temporal attention operations. We consider two U-Net variants of different capacities and a smaller variant of our FIT to evaluate the scalability of both architectures. We detail the architectures in \apref{ap:architecture_details} and show results in Tab.~\ref{table:ablation_architecture_scaling}. 

Our 500M parameters FIT trains 3.31$\times$ faster than the baseline 284M parameters U-Net, performs inference 4.49$\times$ faster and surpasses it in terms of FID and CLIPSIM. In addition, both FITs and U-Nets show strong performance gains with scaling. Our largest FIT scales to 3.9B parameters with only a 1.24$\times$ increase in inference time with respect to the 284M U-Net.

To evaluate the choices operated on our diffusion framework, we ablate different configurations of the diffusion process using our 500M FIT architecture. We produce the following variations: (i) the original EDM framework, (ii) our scaled diffusion framework with EDM $\sigmadata$, (iii) our framework with a reduced value of $\sigmain$, (iv) our framework with images not treated as infinite-frame-rate videos. Our framework improves over EDM under all metrics (i) and shows benefits in setting $\sigmadata=1$, an effect that we attribute to the creation of a training target and loss weighting matching the widely used $\vv$-prediction formulation of \emph{Salimans} \etal \cite{salimans2022progressive} (see Tab.~\ref{table:diffusion_process_comparison}). Using $\sigmain<\upsamplingfactor \sqrt{\frames}$ (see Sec.~\ref{sec:edm_for_high_resolution_video_generation}) impairs performance (iii). Finally, treating images as infinite-frame-rate videos consistently improves FID.

\begin{table}
\begin{center}

\footnotesize
\begin{tabular}{lccc}
\toprule
 & FVD $\downarrow$ & FID $\downarrow$ & IS $\uparrow$ \\
\midrule
 CogVideo \cite{hong2022cogvideo} (Chinese) & 751.3 & - & 23.55 \\
 CogVideo \cite{hong2022cogvideo} (English) & 701.6 & - & 25.27 \\
 MagicVideo \cite{zhou2023magicvideo} & 655 & - & - \\
 LVDM \cite{he2023latent} & 641.8 & - & - \\
 Video LDM \cite{blattmann2023alignyourlatents} & 550.6 & - & \cellthird33.45 \\
 VideoFactory \cite{wang2023videofactory} & 410.0 & - & - \\
 Make-A-Video \cite{singer2022makeavideo} & 367.2 & - & 33.00 \\
 PYoCo \cite{ge2023preserve} & \cellthird355.2 & - & \cellfirst47.46 \\
\midrule
 \methodname{} ($288\times288$ px) & \cellsecond 260.1 & \cellsecond 39.0 & \cellsecond 38.89 \\
 \methodname{} ($512\times288$ px) & \cellfirst 200.2 & \cellfirst 28.1 & \cellsecond 38.89 \\
\bottomrule
\end{tabular}
\end{center}
\caption{Zero-shot evaluation results on UCF101 \cite{soomro2012ucf}.}
\label{table:evaluation_ucf101}

\end{table}

\begin{table}
\begin{center}

\footnotesize
\begin{tabular}{lccc}
\toprule
 & CLIP-FID $\downarrow$ & FVD $\downarrow$ & CLIPSIM $\uparrow$ \\
\midrule
 NUWA \cite{chenfei2022nuwa} (Chinese) & 47.68 & - & 0.2439 \\
 CogVideo \cite{hong2022cogvideo} (Chinese) & 24.78 & - & 0.2614 \\
 CogVideo \cite{hong2022cogvideo} (English) & 23.59 & - & 0.2631 \\
 MagicVideo \cite{zhou2023magicvideo} & - & \cellthird998 & - \\
 LVDM \cite{he2023latent} & - & - & 0.2381 \\
 Latent-Shift \cite{an2023latentshift} & 15.23 & - & 0.2773 \\
 Video LDM \cite{blattmann2023alignyourlatents} & - & - & \cellthird0.2929 \\
 VideoFactory \cite{wang2023videofactory} & - & - & \cellsecond0.3005 \\
 Make-A-Video \cite{singer2022makeavideo} & 13.17 & - & \cellfirst0.3049 \\
 PYoCo \cite{ge2023preserve} & \cellthird9.73 & - & - \\
\midrule
 \methodname{} ($288\times288$ px) & \cellfirst 8.48 & \cellsecond 110.4 & 0.2793 \\
 \methodname{} ($512\times288$ px) & \cellsecond 9.35 & \cellfirst 104.0 & 0.2793 \\
\bottomrule
\end{tabular}
\end{center}
\caption{Zero-shot evaluation results on MSR-VTT \cite{xu2016msrvtt}.}
\label{table:evaluation_mstvtt}

\end{table}

\begin{figure*}
\includegraphics[width=\textwidth]{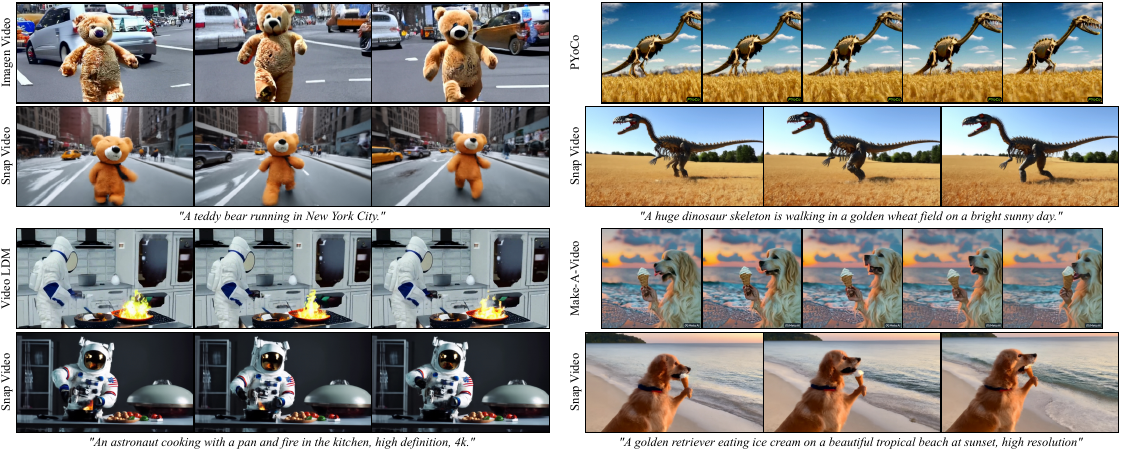}
  \caption{Qualitative results comparing \methodname{} to state-of-the-art video generators on publicly available samples. While baseline methods present motion artifacts (top-left, top-right, bottom-right) or produce \emph{dynamic images} (bottom-left), our method produces more temporally coherent motion. Best viewed in the \website{}.}
  \label{fig:qualitatives_baselines}
\end{figure*}

\subsection{Quantitative Evaluation}
\label{sec:comparison_to_baselines}

We perform comparison of \methodname{} against baselines on the UCF101 \cite{soomro2012ucf}, and MSR-VTT \cite{xu2016msrvtt} datasets respectively in Tab.~\ref{table:evaluation_ucf101} and Tab.~\ref{table:evaluation_mstvtt}. FID and FVD video quality metrics show improvements over the baselines which we attribute to the employed diffusion framework and joint spatiotemporal modeling performed by our architecture. On UCF101, our method produces the second-best IS of $38.89$, demonstrating good video-text alignment. While our method surpasses Make-A-Video \cite{singer2022makeavideo} on UCF101, we note that it produces a lower CLIPSIM score on MSR-VTT. We attribute this behavior to the use of T5 \cite{raffel2022exploring} text embeddings in place of the commonly used CLIP \cite{radford2021clip} embeddings which were observed \cite{saharia2022imagen} to produce higher text-image alignment despite similar CLIPSIM.

\begin{table}
\begin{center}

\setlength{\tabcolsep}{2.0pt}
\footnotesize
\begin{tabular}{lcccc}
\toprule
 & Photorealism & Video-Text Align.& Mot. Quant. & Mot. Qual. \\
\midrule

Gen-2 \cite{esser2023structure} & 44.3 & 81.0 & 96.0 & 78.7 \\
PikaLab \cite{pika} & 61.5 & 80.3 & 89.2 & 70.5  \\
Floor33 \cite{he2023latent} & 76.3 & 80.9 & 88.0 & 79.1 \\
 
\bottomrule
\end{tabular}
\end{center}
\caption{User study on photorealism, video-text alignment, motion quantity and quality against publicly-accessible video generators on 65 dynamic scene prompts. \% of votes in favor of our method.}
\vspace{-2mm}
\label{table:user_study_motion}

\end{table}

To provide a comprehensive evaluation we run a user study to evaluate photorealism, video-text alignment, quantity of motion and quality of motion, important aspects of a video generator. Three publicly-accessible state-of-the-art video generators are considered: Gen-2 \cite{esser2023structure}, PikaLabs \cite{pika} and Floor33 \cite{he2023latent}. We filter a set of 65 prompts from \cite{liu2023evalcrafter} describing scenes with vivid motions, and generate a video for each method with default options. We ask the participants to express preference between paired samples from \methodname{} and each baseline, gathering votes from 5 users for each sample. Results are shown in Tab.~\ref{table:user_study_motion} and video samples provided along with the employed prompt list in \apref{ap:user_studies} and in the \website{}. Our method produces results with photorealism comparable to Gen-2, while surpassing PikaLab and Floor33, and outperforms all baselines with respect to video-text alignment. Most importantly, we note that baselines often produce \emph{dynamic images}, \ie videos with dim motion, or videos with motion artifacts, a finding we attribute to the challenges in modeling large motion. In contrast, our method, thanks to the joint spatiotemporal modeling approach, produces vivid and high-quality motion as shown by the motion metrics.

\subsection{Qualitative Evaluation}
\label{sec:qualitative_evaluation}
In this section, we perform qualitative evaluation of our framework. In Fig.~\ref{fig:qualitatives_baselines}, \apref{ap:qualitatives_against_baselines} and the \website{}, we present qualitative results comparing our method to state-of-the-art generators \cite{ho2022imagenvideo,singer2022makeavideo,blattmann2023alignyourlatents,ge2023preserve} on samples publicly released by the authors. While such prompts might have been selected to highlight strengths of the baselines, our method produces more photorealistic samples aligned to the text descriptions. Most importantly, our samples present vivid and high-quality motion avoiding flickering artifacts that are present in the baselines due to temporal inconsistencies. We accompany qualitative evaluation with a user study performed on the same set of samples in \apref{ap:user_studies}.

\section{Conclusions}
\label{sec:conclusions}

In this work, we highlight the shortcomings of diffusion processes and architectures commonly used in text-to-video generation, and systematically address them by treating videos as first-class citizens. First, we propose a modification to the EDM \cite{karras2022edm} diffusion framework for the generation of high-resolution videos and treat images as high frame-rate videos to avoid image-video modality mismatches. Second, we replace U-Nets \cite{ronneberger2015unet} with efficient transformer-based FITs \cite{chen2023fit} which we scale to billions of parameters. Thanks to their learnable compressed representation of videos, they significantly improve training times, scalability and performance with particular regards to temporal consistency and motion modeling capabilities due to the joint spatiotemporal modeling on the compressed representation. When evaluated on UCF101 \cite{soomro2012ucf} and MSR-VTT \cite{xu2016msrvtt} and in user studies, \methodname{} attains state-of-the-art performance with particular regard to the quality of the modeled motion.

\section{Acknowledgements}
\label{sec:acknowledgements}

We would like to thank Oleksii Popov, Artem Sinitsyn, Anton Kuzmenko, Vitalii Kravchuk, Vadym Hrebennyk, Grygorii Kozhemiak, Tetiana Shcherbakova, Svitlana Harkusha, Oleksandr Yurchak, Andrii Buniakov, Maryna Marienko, Maksym Garkusha, Brett Krong, Anastasiia Bondarchuk
for their help in the realization of video presentations, stories and graphical assets, Colin Eles, Dhritiman Sagar, Vitalii Osykov, Eric Hu for their supporting technical activities, Maryna Diakonova for her assistance with annotation tasks.

{
    \small
    \bibliographystyle{ieeenat_fullname}
    \bibliography{main}
}

\clearpage
\setcounter{page}{1}
\appendix
\maketitlesupplementary


\section{Architecture details}
\label{ap:architecture_details}

In this section, we discuss the details of the architectures employed in this work. Tab.~\ref{table:architecture_details_unet} details the hyperparameters of the UNet \cite{ronneberger2015unet}, while Tab.~\ref{table:architecture_details_fit} shows the hyperparameters of \methodname{} FIT. Note that to ensure divisibility by larger powers of 2, for models producing outputs in $64 \times 36$px resolution we introduce a $4$px vertical padding, yielding a model resolution of $64 \times 40$px. With respect to vanilla FITs \cite{chen2023fit}, our model presents architectural differences that we found beneficial for high-resolution video generation. First, we configure each patch to span over a single frame and each patch group to extend in the temporal dimension to the full extent of the video. We find these modifications to improve temporal modeling. We keep the spatial dimension of the patch to $4 \times 4$px as we found larger patches to degrade spatial modeling capabilities of the model. These settings produce a large number of $147.456$ input patches when processing high-resolution videos. We find the use of local attention layers to be computationally expensive when modeling high-resolution videos, so we replace it with feedforward operations after each cross attention layer. Lastly, our FIT makes use of a large number of latent tokens, a parameter determining the amount of compression applied to the video representation. With respect to typical FIT configurations using no more than $256$ latent tokens, we find it beneficial to increase their number to $768$ when modeling videos to enable additional temporal information to be stored in the latent tokens.

\begin{table}
\begin{center}

\setlength{\tabcolsep}{3.0pt}
\footnotesize
\begin{tabular}{lcc}
    \toprule
    & U-Net 85M & U-Net 284M\\
    \midrule
    Resolution & $16 \times 64 \times 40$ & $16 \times 64 \times 40$ \\
    Base channels & $128$ & $192$ \\
    Channel multiplier & $[1, 2, 2, 3]$ & $[1, 2, 3, 4]$ \\
    Number of residual blocks & $2$ & $2$ \\
    Attention resolutions & $[32, 16, 8]$ & $[32, 16, 8]$ \\
    Attention heads channels & 64 & 64\\
    Conditioning channels & 768 & 768\\
    Label dropout & $0.10$ & $0.10$ \\
    Dropout & $0.10$ & $0.10$ \\
    Use scale shift norm & True & True \\
    \bottomrule
\end{tabular}
\end{center}
\caption{Architectural details of our U-Net baselines.}
\label{table:architecture_details_unet}

\end{table}

\begin{table}
\begin{center}

\setlength{\tabcolsep}{3.0pt}
\footnotesize
\resizebox{\columnwidth}{!}{%
\begin{tabular}{lccc}
    \toprule
    & FIT 500M & FIT 3.9B & FIT 3.9B Up. \\
    \midrule
    Input size & $16 \times 64 \times 40$ & $16 \times 64 \times 40$ & $16 \times 512 \times 288$  \\
    Patch size & $1 \times 4 \times 4$ & $1 \times 4 \times 4$ & $1 \times 4 \times 4$  \\
    Group size & $16 \times 5 \times 4$ & $16 \times 5 \times 4$ & $16 \times 18 \times 16$  \\
    Total patch tokens & $2.560$ & $2.560$ & $147.456$ \\
    Cross att. feedforward & $\checkmark$ & $\checkmark$ & $\checkmark$ \\
    Patch tokens channels & $768$ & $1024$ &  $1024$ \\
    Latent tokens count & $512$ & $768$ &  $768$ \\
    Latent tokens channels & $1024$ & $3072$ &  $3072$ \\
    FIT blocks count & $6$ & $6$ & $6$ \\
    FIT block global layers & $4$ & $4$ & $4$ \\
    FIT block local layers & $0$ & $0$ & $0$ \\
    Patch att. heads channels & $48$ & $64$ &  $64$ \\
    Latent att. heads channels & $64$ & $128$ & $128$  \\
    Self conditioning & $\checkmark$ & $\checkmark$ & $\checkmark$ \\
    Self conditioning prob. & $0.9$ & $0.9$ & $0.9$ \\
    Label dropout & $0.1$ & $0.1$ & $0.1$ \\
    Dropout & $0.1$ & $0.1$ & $0.1$ \\
    Positional embeddings & learnable & learnable & learnable \\
    Conditioning channels & $768$ & $1024$ & $1024$ \\
    \bottomrule
\end{tabular}
}
\end{center}
\caption{Architectural details of \methodname{} FIT models.}
\label{table:architecture_details_fit}

\end{table}

\section{Training details}
\label{ap:training_details}

In this section, we present training details for the models trained in this work, which we summarize in Tab.~\ref{table:training_details}.
We train our model using the LAMB \cite{you2020large} optimizer with a learning rate of $5e^{-3}$ and a cosine learning schedule over 550k steps with 10k warmup steps. To achieve stable training in our largest model we find it important to introduce the following mechanisms. We set $\beta=[0.9, 0.99]$, a weight decay of 0.01, gradient clipping and a total batch size of 2048 videos and 2048 images. We adopt dropout with probability 0.1 in the feedforward transformer layers following self attention operations. Finally, we use an exponential moving average for the model's weights with a base decay halflife of 6k steps.

We implement our model in PyTorch \cite{paszke2019pytorch} and perform all experiments on Nvidia A100 GPUs.

\begin{table}
\begin{center}

\setlength{\tabcolsep}{3.0pt}
\footnotesize
\begin{tabular}{lccc}
    \toprule
    & U-Nets & FIT 3.9B & FIT 3.9B Up. \\
    \midrule
    Optimizer & Adam \cite{kingma2014adam} & LAMB & LAMB \\
    Learning rate & $1e^{-4}$ & $5e^{-3}$ & $5e^{-3}$ \\
    Learning rate decay & cosine & cosine & cosine \\
    Steps & 550k & 550k & 370k \\
    Batch size & $4096$ & $4096$ & $320$ \\
    Samples seen & $2.25$B & $2.25$B & $118$M \\
    Beta & $[0.9, 0.999]$ & $[0.9, 0.99]$ & $[0.9, 0.99]$ \\
    Weight decay & $0.01$ & $0.01$ & $0.01$ \\
    Warmup steps & $10.000$ & $10.000$ & $10.000$ \\
    EMA halflife steps & $6.000$ & $6.000$ & $6.000$ \\
    \bottomrule
\end{tabular}
\end{center}
\caption{Training hyperparameters for our experiments.}
\label{table:training_details}

\end{table}

\section{Additional Evaluation Results and Details}
\label{ap:additional_evaluation_results}

In this section, we provide additional evaluation results for our model. In Sec.~\ref{ap:additional_ablations}, we perform ablations on the sampler parameters. Sec.~\ref{ap:user_studies} presents user study results. In Sec.~\ref{ap:qualitatives_against_baselines} and Sec.~\ref{ap:qualitatives_additional_samples} we show samples against baselines and additional samples from our method. In Sec.~\ref{ap:hierarchical_video_generation}, we describe how to generate longer videos at high framerate. Finally, Sec.~\ref{ap:ucf_101_details} presents evaluation details for UCF101 \cite{soomro2012ucf}.

\subsection{Sampler Parameters Ablations}
\label{ap:additional_ablations}

\begin{figure}
\includegraphics[width=\columnwidth]{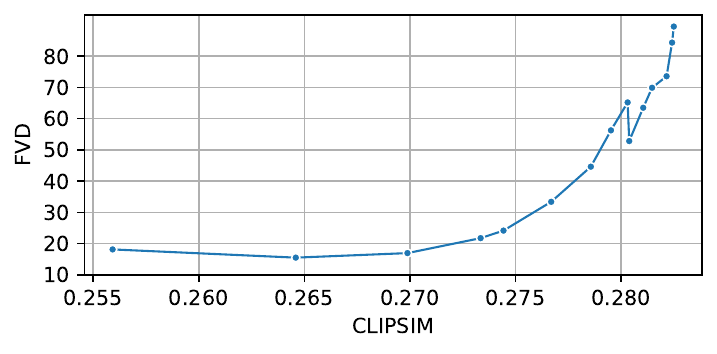}
  \caption{FVD and CLIPSIM on our internal dataset in $64 \times 36$px resolution as a function of the classifier free guidance weight. Points represent weights of $[0,0.5,1,1.5,2,3,4,5,6,7,8,10,12,14,16]$.}
  \label{fig:ablation_cfg_fid_clipscore}
\end{figure}

Choices in the sampling parameters can affect the performance of diffusion models significantly. In Fig.~\ref{fig:ablation_cfg_fid_clipscore}, we show the impact of classifier free guidance \cite{ho2022classifierfree} on model's performance. We produce 10k samples using the first-stage model and report FVD and CLIPSIM at various guidance values, computed on 10k samples with a 40 step deterministic sampler. We find that classifier free guidance can improve both FVD and CLIPSIM, but notice increased sample saturation at high classifier free guidance scales. We adopt dynamic thresholding and oscillating classifier free guidance \cite{saharia2022imagen} which we find effective in reducing the phenomenon.

\begin{figure}
\includegraphics[width=\columnwidth]{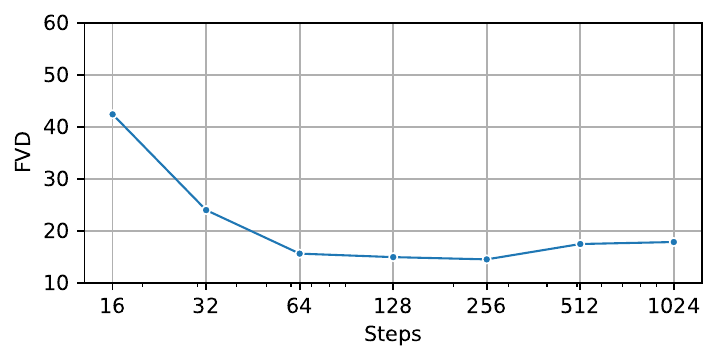}
  \caption{FVD on our internal dataset in $64 \times 36$px resolution as a function of the sampling steps and sampler type.}
  \label{fig:ablation_sampler}
\end{figure}

In Fig.~\ref{fig:ablation_sampler}, we evaluate performance of the model under the same setting, but varying the number of sampling steps. We find that the model produces high-quality samples already at 64 steps and that FVD improves until 256 steps.


\subsection{User Studies}
\label{ap:user_studies}

To complement the evaluation, we run a user study on a set of publicly released samples from Make-A-Video \cite{singer2022makeavideo}, PYoCo \cite{ge2023preserve}, Video LDM \cite{blattmann2023alignyourlatents} and Imagen Video \cite{ho2022imagenvideo}. For each method, we collect publicly available samples and prompts, generate corresponding samples from \methodname{}, and ask participants to express a preference in terms of photorealism, video-text alignment, motion quantity and quality, collecting 5 votes for each sample. Results are shown in Tab.~\ref{table:user_study} and samples are provided in Fig.~\ref{fig:qualitatives_baselines}, Fig.~\ref{fig:qualitatives_baselines_supp} and the \website{}. Our method shows increased photorealism with respect to the baselines, presents higher video-text alignment with respect to all methods except Imagen Video and consistently surpasses baselines in terms of motion quality (see flickering artifacts and temporally inconsistent backgrounds in scenes with large camera motion), a finding we attribute to our joint spatiotemporal modeling approach.

\begin{table}
\begin{center}

\setlength{\tabcolsep}{1.0pt}
\footnotesize
\resizebox{\columnwidth}{!}{%
\begin{tabular}{lcccc}
\toprule
 & Photorealism & Video-Text Align. & Mot. Quant. & Mot. Qual. \\
\midrule

 Imagen Video \cite{ho2022imagenvideo} & 66.9 & 54.3 & 49.4 & 56.3 \\
 PYoCo \cite{ge2023preserve} & 63.3 & 64.9 & 57.1 & 63.9 \\
 Video LDM \cite{blattmann2023alignyourlatents} & 61.7 & 64.4 & 62.2 & 65.8 \\
 Make-A-Video \cite{singer2022makeavideo} & 80.0 & 82.2 & 82.2 & 75.6 \\
 
\bottomrule
\end{tabular}
}
\end{center}
\caption{User study on photorealism, video-text alignment, motion quantity and quality on publicly available samples from closed-source methods. \% of votes in favor of our method.}
\label{table:user_study}

\end{table}

\subsection{Qualitative Results Against Baselines}
\label{ap:qualitatives_against_baselines}

\begin{figure*}
    \includegraphics[width=\textwidth]{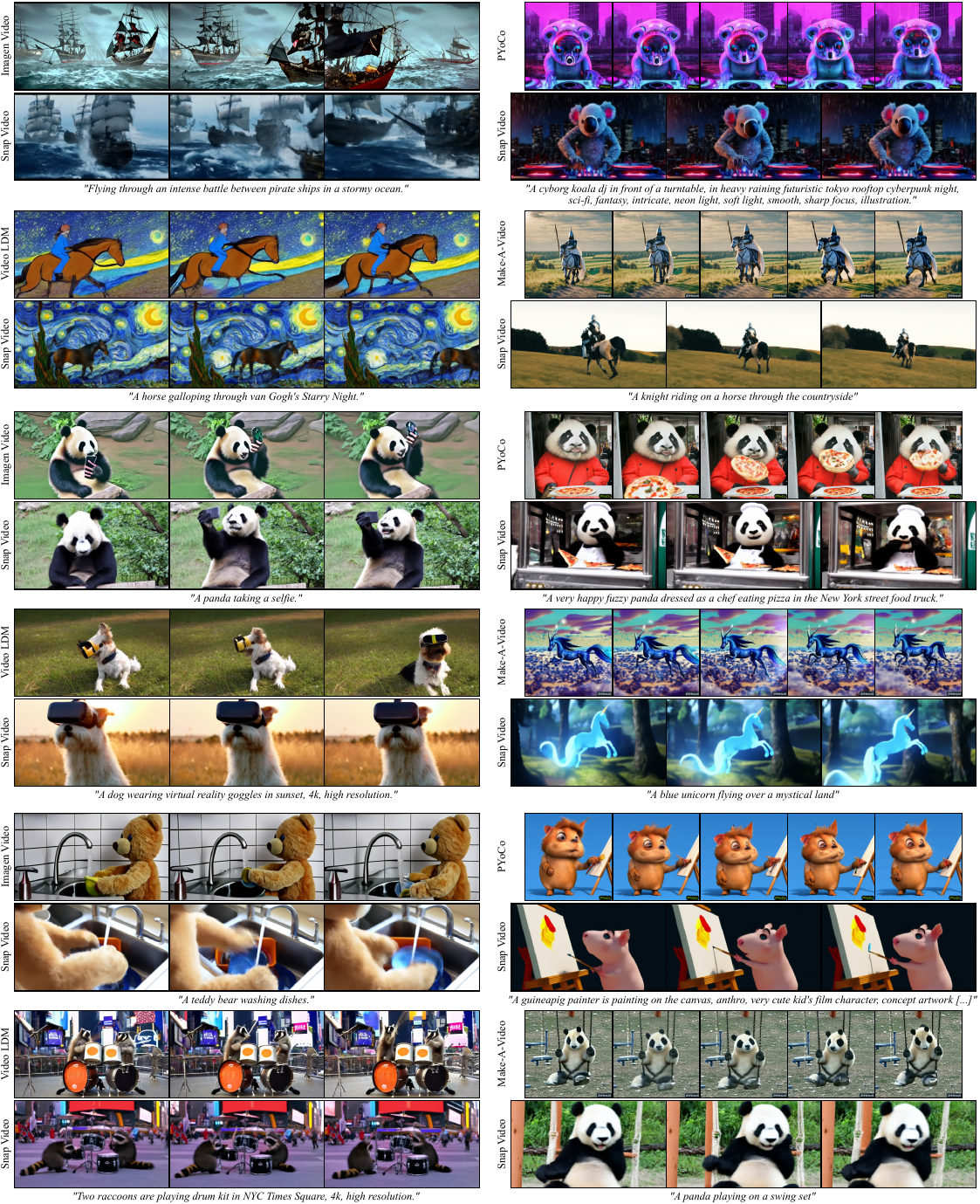}
    \caption{Comparison of \methodname{} to publicly released samples from Make-A-Video \cite{singer2022makeavideo}, PYoCo \cite{ge2023preserve}, Video LDM \cite{blattmann2023alignyourlatents} and Imagen Video \cite{ho2022imagenvideo}. Our method produces temporally coherent motion while avoiding video flickering. Best viewed on the \website{}.}
    \label{fig:qualitatives_baselines_supp}
\end{figure*}

In Fig.~\ref{fig:qualitatives_baselines_supp} and the \website{}, we present qualitative results of our method against baselines on samples publicly released by the authors of Make-A-Video \cite{singer2022makeavideo}, PYoCo \cite{ge2023preserve}, Video LDM \cite{blattmann2023alignyourlatents} and Imagen Video \cite{ho2022imagenvideo}. Our method produces videos with natural motion and can handle scenes with large motion and camera changes while preserving temporal consistency. On the other hand, we observe that baselines often present flickering artifacts and temporally inconsistent objects in case of large motion.

\begin{figure*}
    \includegraphics[width=\textwidth]{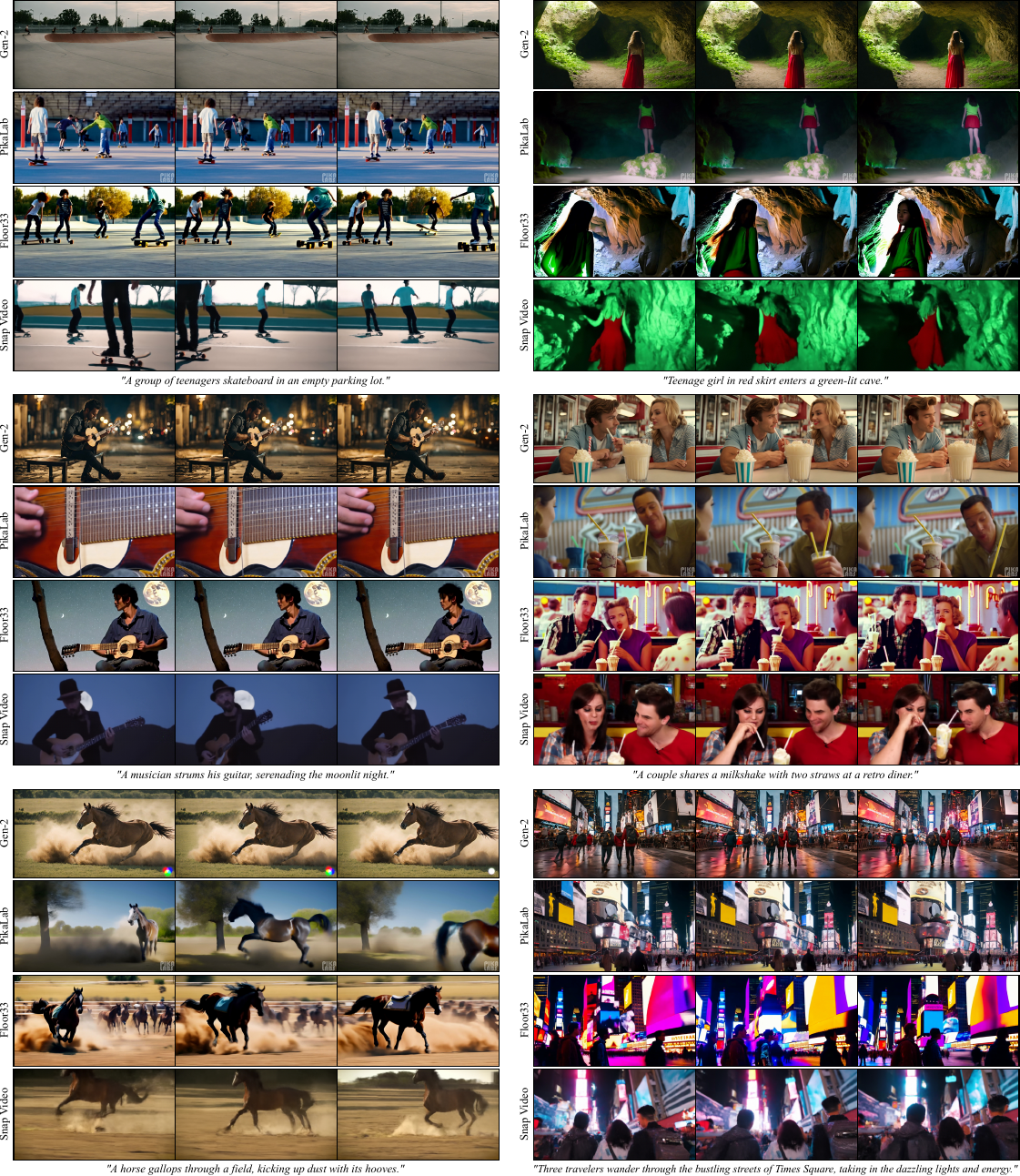}
    \caption{Comparison of \methodname{} to the publicly accessible state-of-the-art video generators Gen-2 \cite{esser2023structure}, PikaLab \cite{pika} and Floor33 \cite{he2023latent}. Rather than producing \emph{dynamic images}, our method generates videos with large amounts of temporally coherent motion. Best viewed on the \website{}.}
    \label{fig:qualitatives_motion_supp}
\end{figure*}

In Fig.~\ref{fig:qualitatives_motion_supp} and the \website{}, we provide qualitative results comparing our method to publicly accessible state-of-the-art video generators including Gen-2 \cite{esser2023structure}, PikaLab \cite{pika} and Floor33 \cite{he2023latent}. Our method produces results that are more aligned to the prompts and, differently from the baselines, which often produce \emph{dynamic images}, our method produces temporally coherent videos with large amounts of motion.

\subsection{Additional Qualitative Results}
\label{ap:qualitatives_additional_samples}

\begin{figure*}
    \includegraphics[width=\textwidth]{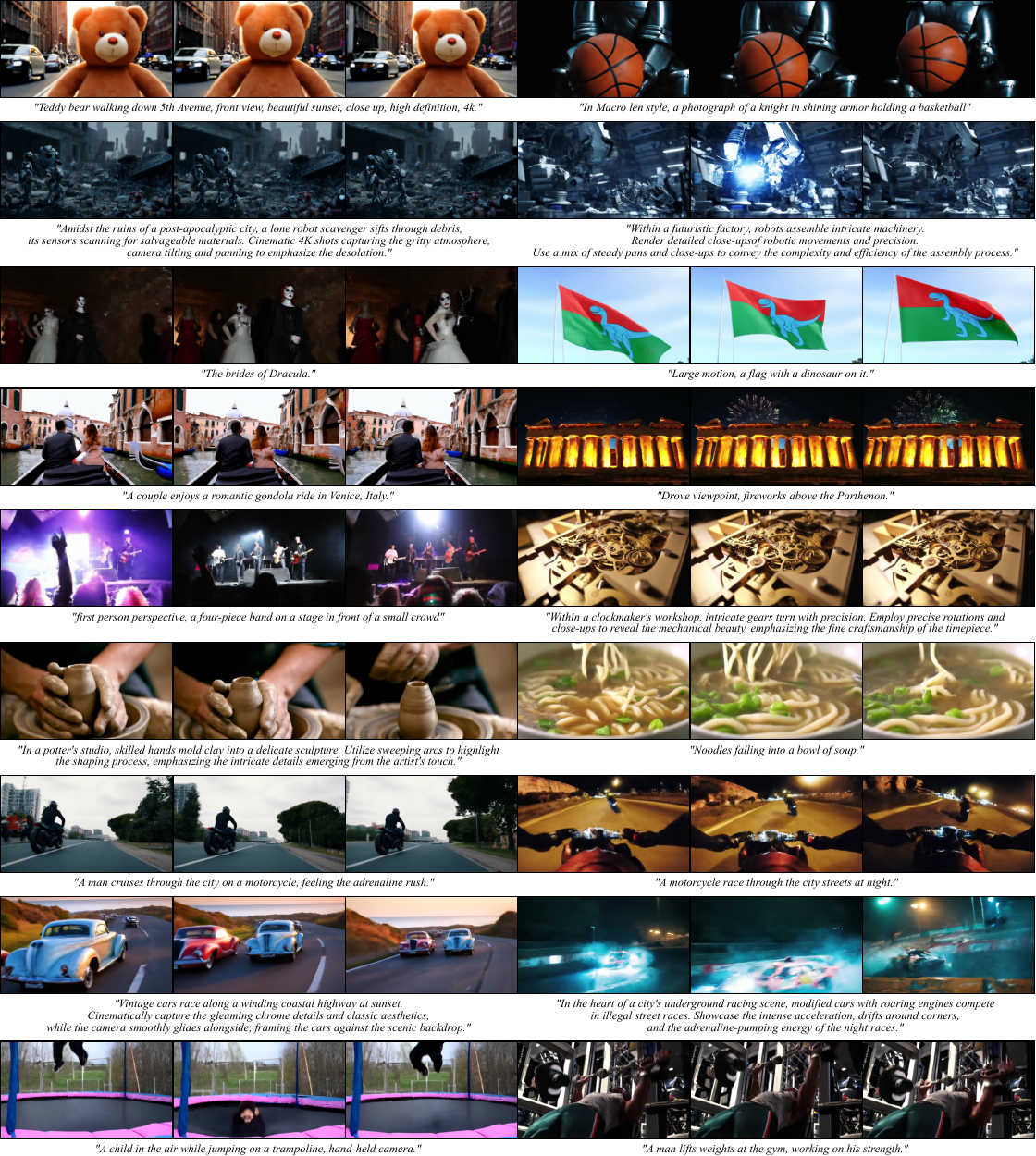}
    \caption{Additional samples generated from \methodname{} on a collection of prompts gathered from ChatGPT, baseline methods and the Evalcrafter \cite{liu2023evalcrafter} benchmark. Best viewed on the \website{}.}
    \label{fig:qualitatives_baselines_p2}
\end{figure*}

\noindent\textbf{Complex Prompts}~We present in Fig.~\ref{fig:qualitatives_baselines_p1}, Fig.~\ref{fig:qualitatives_baselines_p2} and the \website{} additional samples generated by our method on a set of prompts gathered from ChatGPT, Make-A-Video \cite{singer2022makeavideo}, PYoCo \cite{ge2023preserve}, Video LDM \cite{blattmann2023alignyourlatents}, Imagen Video \cite{ho2022imagenvideo}, and the Evalcrafter \cite{liu2023evalcrafter} benchmark. Our method can synthesize a large number of different concepts. Most importantly, it can produce videos with challenging motion including large camera movement, POV videos, and videos of fast-moving objects. Notably, the method maintains temporal consistency and avoids video flickering artifacts. We ascribe these motion modeling capabilities to the joint spatiotemporal modeling performed by our architecture.

\begin{figure*}
    \includegraphics[width=\textwidth]{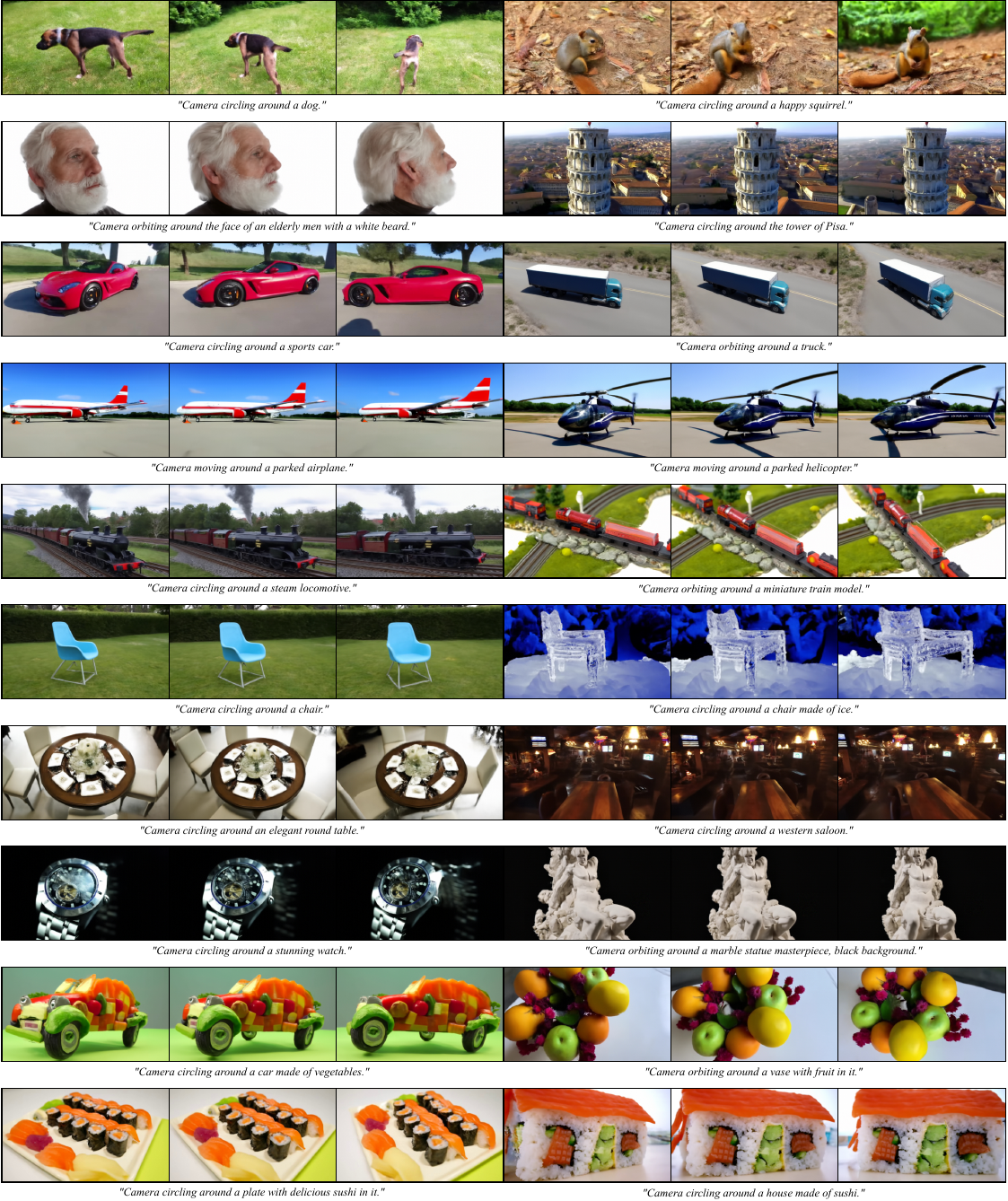}
    \caption{Videos produced by \methodname{} for prompts eliciting circular camera motion around each object. Best viewed on the \website{}.}
    \label{fig:qualitatives_novel_views_supp}
\end{figure*}

\vspace{1.5mm}
\noindent\textbf{Novel Views}~We qualitatively evaluate the capabilities of the proposed method in producing novel views of objects. To do so, we build prompts using one of the following templates \emph{Camera circling around a $\langle$object$\rangle$}, \emph{Camera orbiting around a $\langle$object$\rangle$} or \emph{Camera moving around a $\langle$object$\rangle$}, where \emph{$\langle$object$\rangle$} represents a placeholder for the object to generate. Results are shown in Fig.~\ref{fig:qualitatives_novel_views_supp} and the \website{}. We find that the model is capable of generating plausible novel views of objects, suggesting that it possesses an understanding of the 3D geometry of the modeled objects.

\begin{figure*}
    \includegraphics[width=\textwidth]{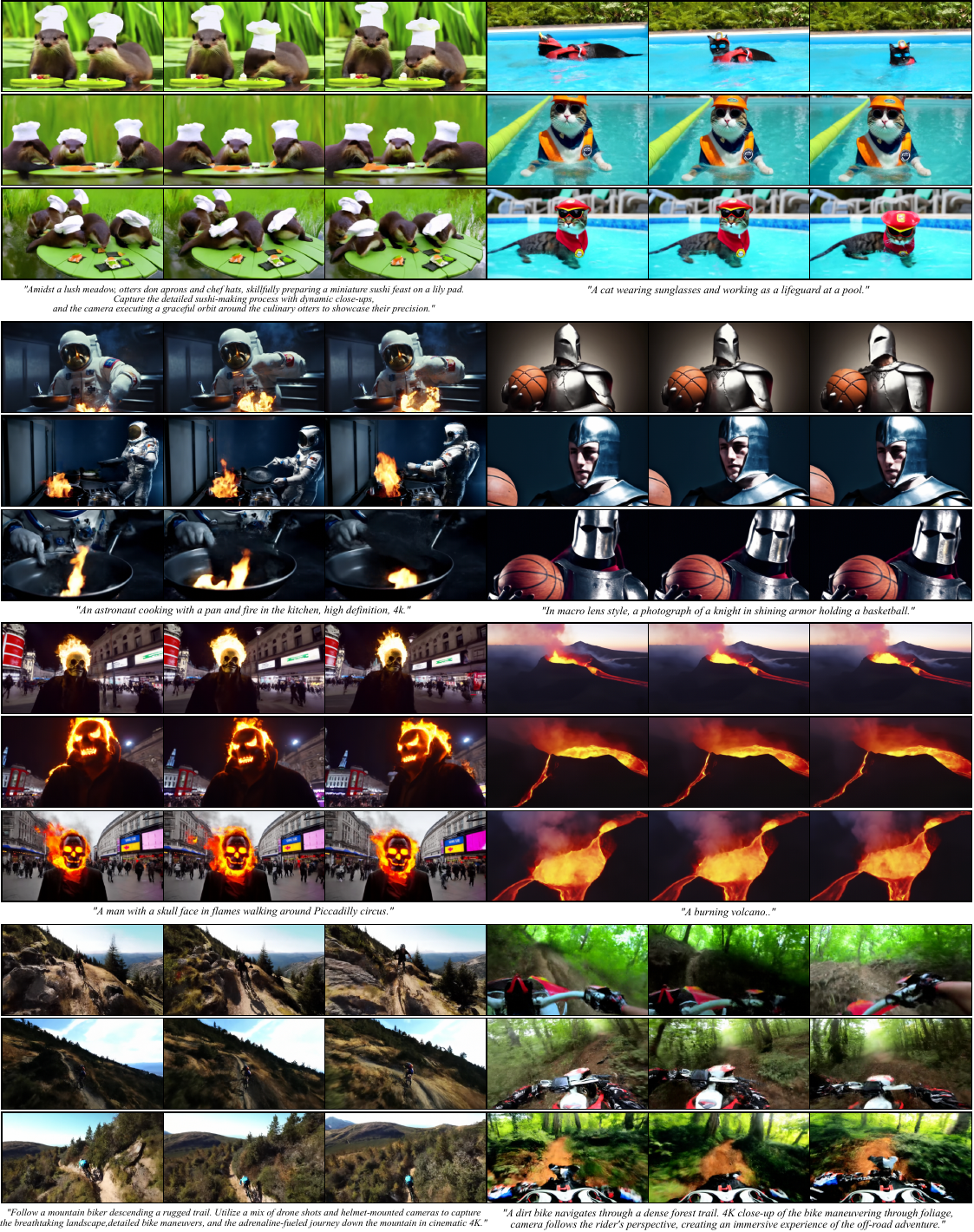}
    \caption{Diversity in samples produced by \methodname{} for the same prompt. See additional samples on the \website{}.}
    \label{fig:qualitatives_diversity_supp}
\end{figure*}

\vspace{1.5mm}
\noindent\textbf{Samples Diversity}~We qualitatively assess the capabilities of the method of generating diverse samples for each prompt. In Fig.~\ref{fig:qualitatives_diversity_supp} and the \website{} we show three samples produced for a set of prompts and note that the method is capable of producing diverse samples.

\begin{figure*}
    \includegraphics[width=\textwidth]{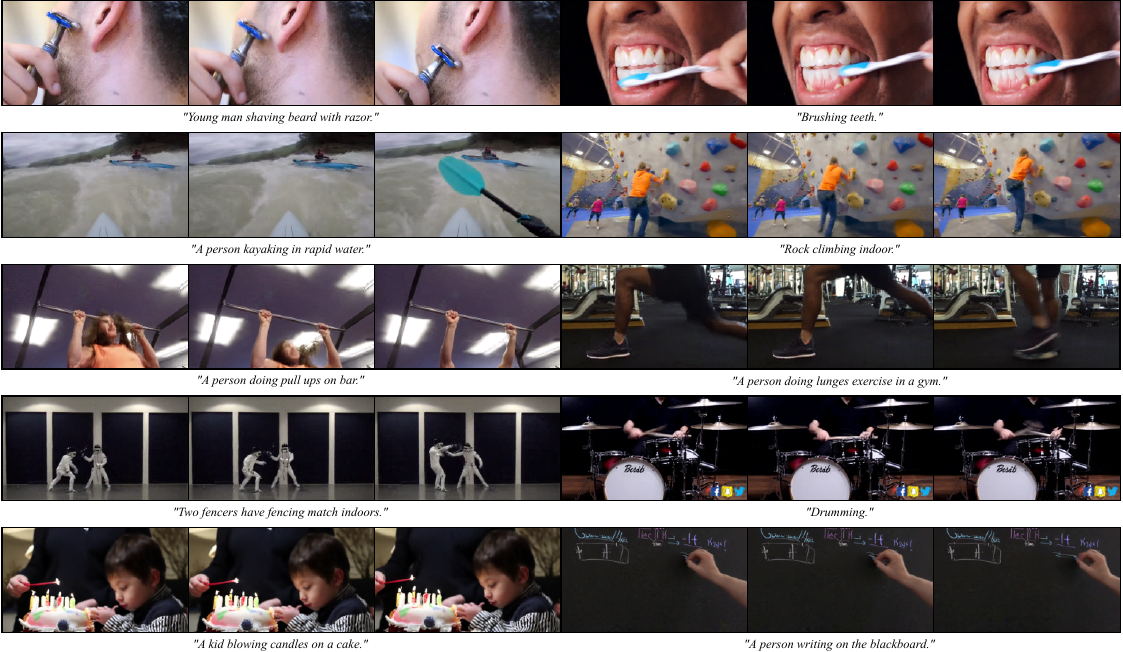}
    \caption{Samples generated by \methodname{} for zero-shot evaluation on UCF101 \cite{soomro2012ucf}. Best viewed on the \website{}.}
    \label{fig:qualitatives_ucf101_supp}
\end{figure*}

\vspace{1.5mm}
\noindent\textbf{UCF 101}~In Fig.~\ref{fig:qualitatives_ucf101_supp} and the \website{}, we present qualitative results produced by our mehthod for zero-shot UCF101 \cite{soomro2012ucf} evaluation.

\subsection{Hierarchical Video Generation}
\label{ap:hierarchical_video_generation}

We train our method to generate videos with a fixed number of frames and variable framerate. We exploit this characteristic and devise a hierarchical generation strategy to increase video duration and framerate by conditioning generation on previously generated frames. In particular, to condition generation on already available frames, we adopt the \emph{reconstruction guidance} method of \emph{Ho} \etal \cite{ho2022video}. We define a hierarchy of progressively increasing framerates and start by autoregressively generating a video of the desired length at the lowest framerate, at each step using the last generated frame as the conditioning. Subsequently, for each successive framerate in the hierarchy, we autoregressively generate a video of the same length but conditioning the model on all frames that have already been generated at the lower framerates. We show samples in the \website{}.

\subsection{Zero-Shot UCF101 Evaluation}
\label{ap:ucf_101_details}

UFC101 \cite{soomro2012ucf} is a dataset of low-resolution Youtube videos. To better match our generated outputs to its distribution, we condition the model to produce videos with a low original resolution through the resolution conditioning mechanism (see Sec.~\ref{sec:architecture}). 
Following \cite{ge2023preserve}, since UCF101 class labels do not always have sufficiently descriptive names, we produce a prompt for each class which we report in the following:
\emph{applying eye makeup}, 
\emph{applying lipstick}, 
\emph{a person shooting with a bow}, 
\emph{baby crawling}, 
\emph{gymnast performing on a balance beam}, 
\emph{band marching}, 
\emph{baseball pitcher throwing baseball}, 
\emph{a basketball player shooting basketball}, 
\emph{dunking basketball in a basketball match}, 
\emph{bench press in a gym}, 
\emph{a person riding a bicycle}, 
\emph{billiards}, 
\emph{a woman using a hair dryer}, 
\emph{a kid blowing candles on a cake}, 
\emph{body weight squats}, 
\emph{a person bowling on bowling alley}, 
\emph{boxing punching bag}, 
\emph{boxing training on speed bag}, 
\emph{swimmer doing breast stroke}, 
\emph{brushing teeth}, 
\emph{a person doing clean and jerk in a gym}, 
\emph{cliff diving}, 
\emph{bowling in cricket match}, 
\emph{batting in cricket match}, 
\emph{cutting in kitchen}, 
\emph{diver diving into a swimming pool from a springboard}, 
\emph{drumming}, 
\emph{two fencers have fencing match indoors}, 
\emph{field hockey match}, 
\emph{gymnast performing on the floor}, 
\emph{group of people playing frisbee on the playground}, 
\emph{swimmer doing front crawl}, 
\emph{golfer swings and strikes the ball}, 
\emph{haircuting}, 
\emph{a person hammering a nail}, 
\emph{an athlete performing the hammer throw}, 
\emph{an athlete doing handstand push up}, 
\emph{an athlete doing handstand walking}, 
\emph{massagist doing head massage to man}, 
\emph{an athlete doing high jump}, 
\emph{group of people racing horse}, 
\emph{person riding a horse}, 
\emph{a woman doing hula hoop}, 
\emph{man and woman dancing on the ice}, 
\emph{athlete practicing javelin throw}, 
\emph{a person juggling with balls}, 
\emph{a young person doing jumping jacks}, 
\emph{a person skipping with jump rope}, 
\emph{a person kayaking in rapid water}, 
\emph{knitting}, 
\emph{an athlete doing long jump}, 
\emph{a person doing lunges exercise in a gym}, 
\emph{a group of soldiers marching in a parade}, 
\emph{mixing in the kitchen}, 
\emph{mopping floor}, 
\emph{a person practicing nunchuck}, 
\emph{gymnast performing on parallel bars}, 
\emph{a person tossing pizza dough}, 
\emph{a musician playing the cello in a room}, 
\emph{a musician playing the daf drum}, 
\emph{a musician playing the indian dhol}, 
\emph{a musician playing the flute}, 
\emph{a musician playing the guitar}, 
\emph{a musician playing the piano}, 
\emph{a musician playing the sitar}, 
\emph{a musician playing the tabla drum}, 
\emph{a musician playing the violin}, 
\emph{an athlete jumps over the bar}, 
\emph{gymnast performing pommel horse exercise}, 
\emph{a person doing pull ups on bar}, 
\emph{boxing match}, 
\emph{push ups}, 
\emph{group of people rafting on fast moving river}, 
\emph{rock climbing indoor}, 
\emph{a person lifting on a rope in a gym}, 
\emph{several people rowing a boat on the river}, 
\emph{a man and a woman are salsa dancing}, 
\emph{young man shaving beard with razor}, 
\emph{an athlete practicing shot put throw}, 
\emph{a teenager skateboarding}, 
\emph{skier skiing down}, 
\emph{jet ski on the water}, 
\emph{a person is skydiving in the sky}, 
\emph{soccer player juggling football}, 
\emph{soccer player doing penalty kick in a soccer match}, 
\emph{gymnast performing on still rings}, 
\emph{sumo wrestling}, 
\emph{surfing}, 
\emph{kids swing at the park}, 
\emph{a person playing table tennis}, 
\emph{a person doing TaiChi}, 
\emph{a person playing tennis}, 
\emph{an athlete practicing discus throw}, 
\emph{trampoline jumping}, 
\emph{typing on computer keyboard}, 
\emph{a gymnast performing on the uneven bars}, 
\emph{people playing volleyball}, 
\emph{walking with dog}, 
\emph{a person standing doing pushups on the wall}, 
\emph{a person writing on the blackboard}, 
\emph{a person at a Yo-Yo competition}.

\begin{figure*}
    \includegraphics[width=\textwidth]{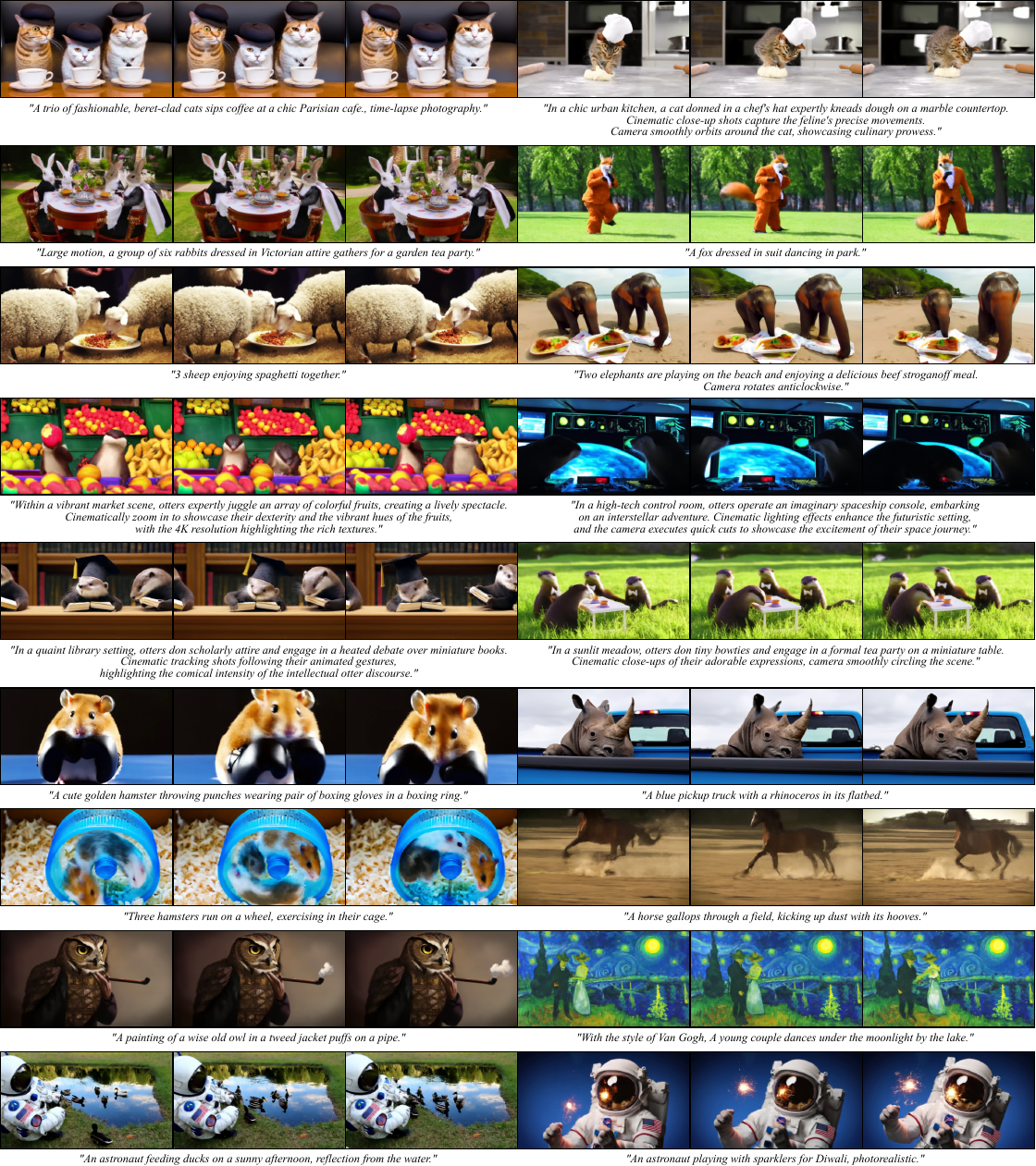}
    \caption{Additional samples generated from \methodname{} on a collection of prompts gathered from ChatGPT, baseline methods and the Evalcrafter \cite{liu2023evalcrafter} benchmark. Best viewed on the \website{}.}
    \label{fig:qualitatives_baselines_p1}
\end{figure*}
\onecolumn
\section{Derivation of EDM Denoising Objective}
\label{ap:edm_derivation}

We derive the denoising objective $\mathcal{L}(\netf)$ expressed in terms of $\netf$ for the EDM framework where we modify the forward process introducing the input scaling factor $\sigmain$:
\begin{eqnarray}
\mathcal{L}(\netf) = \mathbb{E}_{\sigmanoise, \xx, \noise} \Big[ \effectivelossweight(\sigmanoise) ~\big\lVert \netf(\cin(\sigmanoise) \xx_{\sigmanoise}) - \cnorm(\sigmanoise) \target \big\rVert^2_2 \Big]
\label{eq:netf_loss_edm_derivation}
\text{,}
\end{eqnarray}

We start from the denoising objective $\mathcal{L}(\netd)$ as in the original formulation:
\begin{eqnarray}
  \mathcal{L}(\netd) &=& \mathbb{E}_{\sigmanoise, \xx, \noise} \Big[ \lambda(\sigmanoise) ~\big\lVert \netd(\frac{\xx}{\sigmain} + \sigmanoise \noise) - \xx \big\rVert^2_2 \Big]
  \text{,}
\end{eqnarray}

Where we recall the definition of $\netd$:
\begin{equation}
    \netd(\xx_{\sigmanoise}) = \cout(\sigmanoise) \netf\left(\cin(\sigmanoise) \xx_{\sigmanoise}\right) + \cskip(\sigmanoise)(\xx_{\sigmanoise})
    \label{eq:netd_edm_derivation}
    \text{.}
\end{equation}

We also recall the definitions of $\cin(\sigmanoise)$, $\cout(\sigmanoise)$, $\cskip(\sigmanoise)$ , $\lossweight(\sigmanoise)$:
\begin{equation}
  \cin(\sigmanoise) = \frac{1}{\sqrt{\sigmanoise^2 + \sigmadata^2}} \text{, } \quad \cout(\sigmanoise) = \frac{\sigmanoise \cdot \sigmadata}{\sqrt{\sigmanoise^2 + \sigmadata^2}} \text{, } \quad \cskip(\sigmanoise) = \frac{\sigmadata^2}{\big( \sigmanoise^2 + \sigmadata^2 \big)} \text{, } \quad \lossweight(\sigmanoise) = \frac{\sigmanoise^2 + \sigmadata^2}{\sigmanoise^2 \sigmadata^2} \text{. }
\end{equation}

Inserting the definition of $\netd$ and of $\cin(\sigmanoise)$, $\cout(\sigmanoise)$, $\cskip(\sigmanoise)$ into Eq.~\eqref{eq:netf_loss_edm_derivation} we obtain:

\begin{eqnarray}
  && \mathbb{E}_{\sigmanoise, \xx, \noise} \Big[ \lambda(\sigmanoise) \big\lVert \cskip(\sigmanoise) (\frac{\xx}{\sigmain} {+} \sigmanoise \noise) + \cout(\sigmanoise) \netf\big( \cin(\sigmanoise)(\frac{\xx}{\sigmain} {+} \sigmanoise \noise )\big) - \xx \big\rVert^2_2 \Big] \\
  &=& \mathbb{E}_{\sigmanoise, \xx, \noise} \Big[ \frac{\sigmanoise^2 + \sigmadata^2}{\sigmanoise^2 \sigmadata^2} \big\lVert \frac{\sigmadata^2}{\sigmanoise^2 + \sigmadata^2} \big( \frac{\xx}{\sigmain} {+} \sigmanoise \noise \big) + \frac{\sigmanoise \sigmadata}{\sqrt{\sigmanoise^2 + \sigmadata^2}} \netf\big(\cin(\sigmanoise)( \frac{\xx}{\sigmain} {+} \sigmanoise \noise )\big) - \xx \big\rVert^2_2 \Big] \\
  &=& \mathbb{E}_{\sigmanoise, \xx, \noise} \Big[ 1 \cdot \big\lVert \frac{\sigmadata}{\sigmanoise \sqrt{\sigmanoise^2 + \sigmadata^2}} \big(\frac{\xx}{\sigmain} {+} \sigmanoise \noise \big) + \netf\big( \cin(\sigmanoise)(\frac{\xx}{\sigmain} {+} \sigmanoise \noise )\big) - \frac{\sqrt{\sigmanoise^2 + \sigmadata^2}}{\sigmanoise \sigmadata} \xx \big\rVert^2_2 \Big] \\
  &=& \mathbb{E}_{\sigmanoise, \xx, \noise} \Big[ 1 \cdot \big\lVert \netf\big( \cin(\sigmanoise)(\frac{\xx}{\sigmain} {+} \sigmanoise \noise )\big) - \frac{\sigmadata}{\sigmanoise \sqrt{\sigmanoise^2 + \sigmadata^2}} (\xx - \frac{\xx}{\sigmain}) - \frac{\sigmanoise \xx - \sigmadata^2 \noise}{\sigmadata \sqrt{\sigmanoise^2 + \sigmadata^2}} \big\rVert^2_2 \Big] \\
  &=& \mathbb{E}_{\sigmanoise, \xx, \noise} \Big[ 1 \cdot \big\lVert \netf\big( \cin(\sigmanoise)(\frac{\xx}{\sigmain} {+} \sigmanoise \noise )\big)  - \frac{ \frac{\sigmadata^2 (\sigmain - 1)}{\sigmain \sigmanoise}\xx +  \sigmanoise \xx - \sigmadata^2 \noise}{\sigmadata \sqrt{\sigmanoise^2 + \sigmadata^2}} \big\rVert^2_2 \Big]
  \text{.}
\end{eqnarray}

From which follows that:
\begin{equation}
\target = \sigmanoise \xx - \sigmadata^2 \noise + \frac{\sigmadata^2 (\sigmain - 1)}{\sigmain \sigmanoise} \xx \text{, } \quad \cnorm(\sigmanoise) = \frac{1}{\sigmadata \sqrt{\sigmanoise^2 + \sigmadata^2}} \text{, } \quad \effectivelossweight(\sigmanoise) = 1 \text{. }
\end{equation}

Note that the training target has a spurious term $\frac{\sigmadata^2 (\sigmain - 1)}{\sigmain \sigmanoise}\xx$ which approaches infinity as $\sigmanoise$ approaches 0.

From this formulation we also notice the link between EDM and the $\vv$-prediciton framework. First, the training target consists in a rescaled and negated $\vv$-prediction objective with $\vv=\sigmadata^2 \noise - \sigmanoise \xx$. Second the loss weight equals to a reweighted $1 + \snr$  $\vv$-prediction framework \cite{salimans2022progressive} weighting:
\begin{eqnarray}
  \lossweight(\sigmanoise) &=& \frac{\sigmanoise^2 + \sigmadata^2}{\big(\sigmanoise \sigmadata\big)^2} = \frac{1}{\sigmadata^2} + \frac{1}{\sigmanoise^2} = \frac{1}{\sigmadata^2} + \snr \text{. }
\end{eqnarray}
Thus, when $\sigmadata=1$, EDM is equivalent to the $\vv$-prediciton formulation. Starting from these observations, we rewrite the framework so that it exhibits a well-formed $\target$ for all values of $\sigmanoise$ by avoiding the spurious term $\frac{\sigmadata^2 (\sigmain - 1)}{\sigmain \sigmanoise}\xx$.

\section{Derivation of Our Diffusion Framework}
\label{ap:our_framework_derivation}

We start the derivation of our diffusion framework by imposing that the training target equals the original EDM $\vv$-prediction objective for all values of $\sigmain$, without the spurious term $\frac{\sigmadata^2 (\sigmain - 1)}{\sigmain \sigmanoise}\xx$ affecting the original formulation for $\sigmain \neq 1$ (see \apref{ap:edm_derivation}).
\begin{equation}
  \target = \vv = \sigmadata^2 \noise - \sigmanoise \xx
  \text{.}
\end{equation}
We then derive $\cnorm(\sigmanoise)$ such that $\cnorm(\sigmanoise) \target$, the function approximated by $\netf$, has unit variance:
\begin{eqnarray}
  \Var_{\xx, \noise} \big[ \cnorm(\sigmanoise) \target \big] &=& 1 \\
  \Var_{\xx, \noise} \Big[ \cnorm(\sigmanoise) \big( \sigmadata^2 \noise - \sigmanoise \xx \big) \Big] &=& 1 \\
  \cnorm(\sigmanoise)^2 &=& \frac{1}{\Var_{\xx, \noise} \Big[ \big( \sigmadata^2 \noise - \sigmanoise \xx \big) \Big]} \\
  \cnorm(\sigmanoise)^2 &=& \frac{1}{\sigmadata^2 (\sigmadata^2 + \sigmanoise^2)} \\
  \cnorm(\sigmanoise) &=& \frac{1}{\sigmadata \sqrt{ (\sigmadata^2 + \sigmanoise^2)}} \text{.}
\end{eqnarray}
Following standard normalization practices, we define $\cin(\sigmanoise)$ so that the model input has unit variance:
\begin{eqnarray}
  \Var_{\xx, \noise} \big[ \cin(\sigmanoise) (\frac{\xx}{\sigmain} + \sigmanoise \noise) \big] &=& 1 \\
  \cin(\sigmanoise)^2 &=& \frac{1}{\Var_{\xx, \noise} \big[ \frac{\xx}{\sigmain} + \sigmanoise \noise \big]} \\
  \cin(\sigmanoise)^2 &=& \frac{1}{\frac{\sigmadata^2}{\sigmain^2} + \sigmanoise^2}\\
  \cin(\sigmanoise) &=& \frac{1}{\sqrt{\frac{\sigmadata^2}{\sigmain^2} + \sigmanoise^2}} \text{.}
\end{eqnarray}

To derive the remaining framework components, we first recall the definition of our forward process:
\begin{equation}
  \xx_{\sigmanoise} = \frac{\xx}{\sigmain} + \sigmanoise \noise
  \text{,}
\end{equation}
and note that $\xx$ can be recovered from $\xx_{\sigmanoise}$ and $\vv$ as:
\begin{equation}
  \xx = \frac{\xx_{\sigmanoise} - \frac{\sigmanoise}{\sigmadata^2}\vv}{\frac{1}{\sigmain} + \frac{\sigmanoise^2}{\sigmadata^2}} \label{eq:x_from_v}
  \text{,}
\end{equation}
and consequently
\begin{equation}
  \vv = \frac{\sigmadata^2 \xx_{\sigmanoise} - (\frac{\sigmadata^2}{\sigmain} + \sigmanoise^2) \xx}{\sigmanoise} \label{eq:v_from_xsigma}
  \text{.}
\end{equation}
To recover $\cskip(\sigmanoise)$ and $\cout(\sigmanoise)$ we note from the definition of $\target = \vv$ and the loss expressed in Eq.~\eqref{eq:netf_loss_edm_derivation} that as it approaches zero the following holds:
\begin{eqnarray}
    \cnorm(\sigmanoise) \target &=&  \netf(\cin(\sigmanoise) \xx_{\sigmanoise})\\
    \vv &=& \frac{\netf(\cin(\sigmanoise) \xx_{\sigmanoise})}{\cnorm(\sigmanoise)} \\
    \vv &=& \sigmadata \sqrt{\sigmanoise^2 + \sigmadata^2} \netf(\cin(\sigmanoise) \xx_{\sigmanoise}) \label{eq:v_as_ftarget} \text{.}
\end{eqnarray}
Substituting Eq.~\eqref{eq:v_as_ftarget} into Eq.~\eqref{eq:x_from_v} we obtain:
\begin{eqnarray}
    \netd(\xx_{\sigmanoise}) = \xx &=& \frac{\xx_{\sigmanoise} - \frac{\sigmanoise}{\sigmadata^2}\vv}{\frac{1}{\sigmain} + \frac{\sigmanoise^2}{\sigmadata^2}} \\
  &=& \frac{\xx_{\sigmanoise}}{\frac{1}{\sigmain} + \frac{\sigmanoise^2}{\sigmadata^2}} - \frac{\frac{\sigmanoise}{\sigmadata^2}\vv}{\frac{1}{\sigmain} + \frac{\sigmanoise^2}{\sigmadata^2}} \\
  &=& \frac{\xx_{\sigmanoise}}{\frac{1}{\sigmain} + \frac{\sigmanoise^2}{\sigmadata^2}} - \frac{\frac{\sigmanoise}{\sigmadata} \sqrt{\sigmanoise^2 + \sigmadata^2} \netf(\cin(\sigmanoise) \xx_{\sigmanoise})}{\frac{1}{\sigmain} + \frac{\sigmanoise^2}{\sigmadata^2}} \\
  &=& \cskip(\sigmanoise) \xx_{\sigmanoise} + \cout(\sigmanoise)\netf(\cin(\sigmanoise) \xx_{\sigmanoise})
  \text{,}
\end{eqnarray}
from which we recognize:
\begin{equation}
  \cskip(\sigmanoise) = \frac{\sigmain\sigmadata^2}{\sigmain\sigmanoise^2 + \sigmadata^2} \text{, } \quad \cout(\sigmanoise) = -\sigmain\sigmanoise \sigmadata \frac{ \sqrt{\sigmanoise^2 + \sigmadata^2}}{\sigmadata^2 + \sigmain\sigmanoise^2} \text{.}
\end{equation}
We set the loss weight $\lossweight(\sigmanoise)$ to the same value as EDM:
\begin{equation}
\lossweight(\sigmanoise) = \frac{1}{\sigmadata^2} + \frac{1}{\sigmanoise^2} \text{.}
\end{equation}
To recover $\effectivelossweight(\sigmanoise)$, inserting the definition of $\netd$ (Eq.~\eqref{eq:netd_edm_derivation}) and of $\cin(\sigmanoise)$, $\cout(\sigmanoise)$, $\cskip(\sigmanoise)$, $\lossweight(\sigmanoise)$ into Eq.~\eqref{eq:netf_loss_edm_derivation} we obtain:
\begin{eqnarray}
  && \mathbb{E}_{\sigmanoise, \xx, \noise} \Big[ \lambda(\sigmanoise) \big\lVert \cskip(\sigmanoise) (\frac{\xx}{\sigmain} {+} \sigmanoise \noise) + \cout(\sigmanoise) \netf\big( \cin(\sigmanoise)\xx_{\sigmanoise} \big) - \xx \big\rVert^2_2 \Big] \\
  &=& \mathbb{E}_{\sigmanoise, \xx, \noise} \Big[ \lambda(\sigmanoise) \big\lVert \frac{\sigmain\sigmadata^2}{\sigmain\sigmanoise^2 + \sigmadata^2} (\frac{\xx}{\sigmain} {+} \sigmanoise \noise) -\sigmain\sigmanoise \sigmadata \frac{ \sqrt{\sigmanoise^2 + \sigmadata^2}}{\sigmadata^2 + \sigmain\sigmanoise^2} \netf\big( \cin(\sigmanoise)\xx_{\sigmanoise} \big) - \xx \big\rVert^2_2 \Big] \\
  &=& \mathbb{E}_{\sigmanoise, \xx, \noise} \Big[ \frac{\lambda(\sigmanoise)}{(\sigmanoise^2 + \frac{\sigmadata^2}{\sigmain})^2} \big\lVert  \frac{\sigmanoise}{\cnorm(\sigmanoise)}\netf\big( \cin(\sigmanoise)\xx_{\sigmanoise} \big) + \sigmanoise^2 \xx - \sigmadata^2 \sigmanoise \noise \big\rVert^2_2 \Big] \\
    &=& \mathbb{E}_{\sigmanoise, \xx, \noise} \Big[ \frac{\sigmanoise^2}{(\sigmanoise^2 + \frac{\sigmadata^2}{\sigmain})^2} \frac{\lambda(\sigmanoise)}{\cnorm(\sigmanoise)^2} \big\lVert  \netf\big(\cin(\sigmanoise)\xx_{\sigmanoise}\big) + \cnorm(\sigmanoise)(\sigmanoise \xx - \sigmadata^2 \noise) \big\rVert^2_2 \Big] \\
    &=& \mathbb{E}_{\sigmanoise, \xx, \noise} \Big[ \frac{(\sigmanoise^2 + \sigmadata^2)^2}{(\sigmanoise^2 + \frac{\sigmadata^2}{\sigmain})^2} \big\lVert  \netf\big( \cin(\sigmanoise)\xx_{\sigmanoise} \big) + \cnorm(\sigmanoise)\target \big\rVert^2_2 \Big] \\
    &=& \mathbb{E}_{\sigmanoise, \xx, \noise} \Big[ \effectivelossweight(\sigmanoise) \big\lVert  \netf\big( \cin(\sigmanoise)\xx_{\sigmanoise}  \big) + \cnorm(\sigmanoise)\target \big\rVert^2_2  \Big] \text{,}
\end{eqnarray}
from which:
\begin{equation}
  \effectivelossweight(\sigmanoise) = \frac{(\sigmanoise^2 + \sigmadata^2)^2}{(\sigmanoise^2 + \frac{\sigmadata^2}{\sigmain})^2} \text{.}
\end{equation}

In conclusion, the proposed diffusion framework maintains the training target $\target$ equal to the original EDM training target for all values of the input scaling factor $\sigmain$, ensuring that $\netf$ learns the same denoising function while giving control on the quantity of the signal present in the input. In addition, our framework preserves the original loss weight $\lossweight(\sigmanoise)$, giving control over the input scaling factor without affecting the weight of the loss over the different noise levels.
\twocolumn
\section{Discussion}
\label{ap:discussion}

In this section, we describe the limitations of our framework (see Sec.~\ref{ap:limitations}) and discuss societal impact (see Sec.~\ref{ap:ethical_considerations}).

\subsection{Limitations}
\label{ap:limitations}

\methodname{} presents limitations which we discuss in this section.

\noindent\textbf{Text Rendering} We find our framework to often spell text incorrectly. We attribute this finding to a lack of high-quality videos depicting text matched by the exact description of the displayed text. Automated pipelines for OCR can be employed on the training dataset to address this issue.

\noindent\textbf{Object count} Similarly, the generator may not render the requested number of entities, especially when the cardinality of the objects is high. The behavior can be explained by the difficulties in learning correct object counts from video data, where descriptions can be noisy, objects can enter and exit the scene and the camera can move widely, changing the cardinality of objects in each frame.

\noindent\textbf{Positional understanding} While the generator can place objects in positions requested by the prompts, we find it can not reliably synthesize videos corresponding to prompts entailing complex positional relationships between multiple entities such as \emph{``A stack of three cubes: the top one is blue, the bottom one green and the middle one is red''}. 

\noindent\textbf{Stylization} We find that our model can generate stylized content (see Fig.~\ref{fig:qualitatives_baselines_p1}), but may present failure cases where the style specification is ignored or the stylized contents only translate in the scene rather than being animated. We attribute this finding to the lack of model training on a filtered set of data presenting high aesthetic scores which we find to contain textual descriptions related to artistic and visual styles with higher probability.

\noindent\textbf{Negation} Some challenging prompts such as \emph{``A glass of juice next to a plate with no bananas in it''} may lead the model to ignore the negation, resulting in the generation of all entities.

\noindent\textbf{Block artifacts} Videos may contain content with a very large amount of motion, leading to a greater difficulty in compressing its content to a latent representation of fixed size. In such situations we find that the model may produce patch tokens that do not blend together in a perfect manner, resulting in some visible patches in the videos, akin to video compression artifacts.

\noindent\textbf{Resolution} Our two-stage model cascade generates videos in $512 \times 288$px resolution. We note that generating video content aligning to the given prompt and presenting temporally coherent motion are the most critical problems in video generation, and possible artifacts in these categories are already visible in $512 \times 288$px resolution, as shown in our comparisons to baselines. We also note that cascaded model stages are independently trained and agnostic to the employed previous-stage generator. Thus, given an upsampler from $512 \times 288$px to a higher resolution, any improvement shown in $512 \times 288$px resolution with respect to baselines is expected to produce higher resolution results of correspondingly improved quality. We consider the integration of additional cascade stages as an interesting venue for future work.

\subsection{Societal Impact}
\label{ap:ethical_considerations}

Text-to-video generative models are evolving rapidly \cite{ho2022imagenvideo,ge2023preserve,blattmann2023alignyourlatents,wang2023videofactory} and hold promise to empower users with new and powerful ways to express their creativity once accessible only to trained experts such as artists and digital content creators. With such improvements comes a greater risk that generated results may be perceived as real with the potential for nefarious individuals to generate harmful or deceiving content. Our model is exposed to a broad range of concepts during training and makes use of a T5 \cite{raffel2022exploring} text encoder that was trained on unfiltered internet data, making it necessary to guard it against such possible uses. In addition, the model generates data following its training data distribution which implies that potential biases that may be present in the dataset can be reflected in the model outputs. To avoid misuse, we do not make the model publicly accessible and plan to put in place data cleaning, prompt filtering and output filtering techniques and watermarking as additional safeguards.


\end{document}